\documentclass[sigplan, 10pt, screen, authorversion]{acmart}
\usepackage[T1]{fontenc}
\usepackage[utf8]{inputenc}
\usepackage{booktabs} % For formal tables
\usepackage{color}
\usepackage{balance}
\usepackage{amsfonts}
\usepackage{amssymb, bbm}
\usepackage{multirow, multicol}
\usepackage{caption, subcaption}
\usepackage{comment}
\usepackage{xspace}
\usepackage{upquote}
\usepackage{float}
\usepackage{alltt}
\usepackage{algpseudocode}
\usepackage{pbox}
\usepackage{enumitem}
\usepackage{mathpartir}
\usepackage{wrapfig}
\usepackage{hhline}
\usepackage{stmaryrd}
\usepackage{pifont}
\usepackage{makecell}
\usepackage{graphicx}
\usepackage{amsmath}
\usepackage{array}
\usepackage{ccicons}
\usepackage{listings}
\usepackage{url}

\newcommand{\WHEN}[0]{\textsc{when}\xspace}
\definecolor{darkorange}{RGB}{220,136,8}
\newcommand{\GET}[0]{\textsc{get}\xspace}
\definecolor{darkgreen}{RGB}{34,139,34}
\newcommand{\DO}[0]{\textsc{do}\xspace}

\newcommand{\thingtalk}[0]{ThingTalk\xspace}
\newcommand{\TT}[0]{ThingTalk\xspace}
\newcommand{\TTa}[0]{TT$+$A\xspace}

\newcommand{\TACL}[0]{TACL\xspace}

\newcommand{\NLPL}[0]{VAPL\xspace}
\newcommand{\VAPL}[0]{VAPL\xspace}

\definecolor{comment-red}{rgb}{0.8,0,0}
\newcommand{\TODO}[1]{{\color{red} TODO: #1}}

\newcommand{\INTERVAL}{{\mbox{\small{\textsc{INTERVAL}}}}\xspace}

\newcommand{\COMMAND}{{\mbox{{\textsc{command}}}}\xspace}
\newcommand{\WHPHRASE}{{\mbox{{\textsc{wp}}}}\xspace}
\newcommand{\NOUNPHRASE}{{\mbox{{\textsc{np}}}}\xspace}
\newcommand{\VERBPHRASE}{{\mbox{{\textsc{vp}}}}\xspace}

\copyrightyear{2019}
\acmYear{2019}
\setcopyright{acmlicensed}
\acmConference[PLDI '19]{Proceedings of the 40th ACM SIGPLAN Conference on Programming Language Design and Implementation}{June 22--26, 2019}{Phoenix, AZ, USA}
\acmBooktitle{Proceedings of the 40th ACM SIGPLAN Conference on Programming Language Design and Implementation (PLDI '19), June 22--26, 2019, Phoenix, AZ, USA}
\acmPrice{15.00}
\acmDOI{10.1145/3314221.3314594}
\acmISBN{978-1-4503-6712-7/19/06}

\begin{document}
\title[Genie: A Generator of NL Semantic Parsers for VA Commands]
{Genie: A Generator of Natural Language \\Semantic Parsers for Virtual Assistant Commands}

%\author[Giovanni Campagna, Silei Xu, Mehrad Moradshahi, Monica S. Lam]{Giovanni Campagna\footnote{Equal contribution} \quad Silei Xu* \quad Mehrad Moradshahi \quad Monica S. Lam}
\author[G. Campagna]{Giovanni Campagna}\authornote{Equal contribution}
\affiliation{%
  \institution{Computer Science Department}
  \institution{Stanford University}
  \streetaddress{353 Serra Mall}
  \city{Stanford}
  \state{CA}
  \postcode{94305}
  \country{USA}
}
\email{gcampagn@cs.stanford.edu}
\author[S. Xu]{Silei Xu}\authornotemark[1]
\affiliation{%
  \institution{Computer Science Department}
  \institution{Stanford University}
  \streetaddress{353 Serra Mall}
  \city{Stanford}
  \state{CA}
  \postcode{94305}
  \country{USA}
}
\email{silei@cs.stanford.edu}
\author[M. Moradshahi]{Mehrad Moradshahi}
%\thanks{Equal contribution}
\affiliation{%
  \institution{Computer Science Department}
  \institution{Stanford University}
  \streetaddress{353 Serra Mall}
  \city{Stanford}
  \state{CA}
  \postcode{94305}
  \country{USA}
}
\email{mehrad@cs.stanford.edu}
\author[R. Socher]{Richard Socher}
\affiliation{%
  \institution{Salesforce, Inc.}
  \streetaddress{502 Emerson St}
  \city{Palo Alto}
  \state{CA}
  \postcode{94301}
  \country{USA}
}
\email{rsocher@salesforce.com}
\author[M. S. Lam]{Monica S. Lam}
\affiliation{%
  \institution{Computer Science Department}
  \institution{Stanford University}
  \streetaddress{353 Serra Mall}
  \city{Stanford}
  \state{CA}
  \postcode{94305}
  \country{USA}
}
\email{lam@cs.stanford.edu}

% If the default list of authors is too long for headers.
%\renewcommand{\shortauthors}{}

\begin{abstract}
To understand diverse natural language commands, virtual assistants today are trained with numerous labor-intensive, manually annotated sentences. This paper presents a methodology and the Genie toolkit that can handle new compound commands with significantly less manual effort. 

We advocate formalizing the capability of virtual assistants with a {\em Virtual Assistant Programming Language (\NLPL)} and using a neural semantic parser to translate natural language into VAPL code.  Genie needs only a small realistic set of input sentences for validating the neural model. Developers write templates to synthesize data; Genie uses crowdsourced paraphrases and data augmentation, along with the synthesized data, to train a semantic parser. 
%Developers can tune the templates iteratively until good quality is achieved.

We also propose design principles that make \VAPL languages amenable to natural language translation.  
We apply these principles to revise ThingTalk, the language used by the Almond virtual assistant.  We use Genie to build the first semantic parser that can support compound virtual assistants commands with unquoted free-form parameters.
Genie achieves a 62\% accuracy on realistic user inputs.  We demonstrate Genie's generality by showing a 
19\% and 31\% improvement over the previous state of the art on a music skill, aggregate functions, and access control.

\end{abstract}





%  \caption{This is a teaser}

\maketitle

%!TEX root = ./paper.tex
\begin{figure}[hb]
\vspace{-1.5em}
\centering
\includegraphics[width=\linewidth]{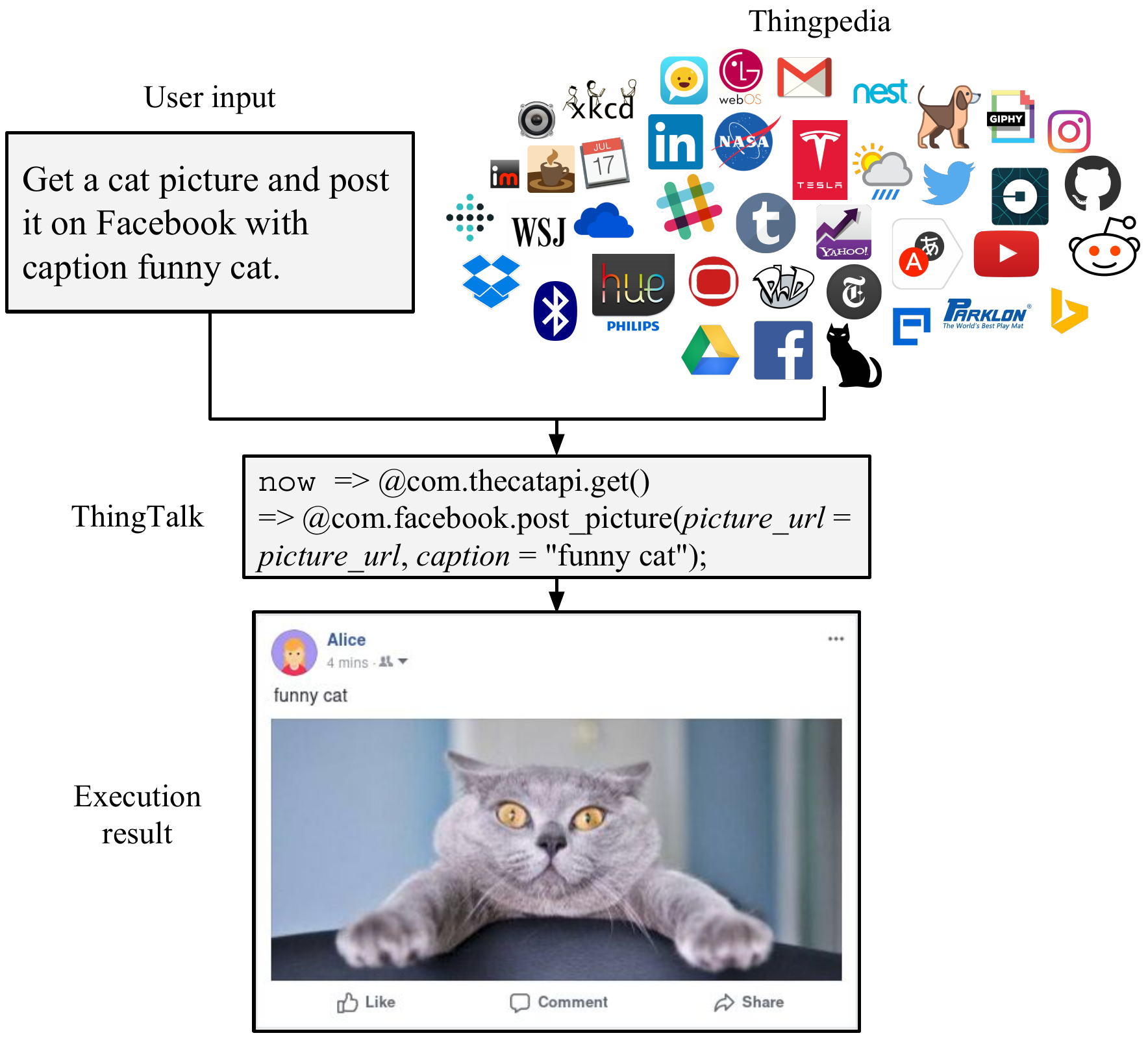}
\caption{An example of translating and executing compound virtual assistant commands.}
\label{fig:running_example}
\vspace{-1em}
\end{figure}
\section{Introduction}
\label{sec:intro}

Personal virtual assistants provide users with a natural language interface to a wide variety of web services and IoT devices. Not only must they understand primitive commands across many domains, but they must also understand the composition of these commands to perform entire tasks.  State-of-the-art virtual assistants are based on \textit{semantic parsing}, a machine learning algorithm that converts natural language to a semantic representation in a formal language.  
The breadth of the virtual assistant interface makes it particularly challenging to design the semantic representation.  Furthermore, there is no existing corpus of natural language commands to train the neural model for new capabilities.  This paper advocates using a {\em Virtual Assistant Programming Language} (VAPL) to capture the formal semantics of the virtual assistant capability.  We also present Genie, a toolkit for creating a semantic parser for new virtual assistant capabilities that can be used to bootstrap real data acquisition.

%They are powerful as they support many different services and domains, but their development and growth to new domains is limited by the need to collect annotated data to train the natural language understanding part. This paper asks how we can bootstrap a new virtual assistant, when there is no existing data? What should be the interface between natural language (NL) and the execution engine of the assistants?  How can we easily extend virtual assistants with new devices and new constructs? More importantly, how can we teach virtual assistants to understand more commands in natural language easily?

\subsection{Virtual Assistant Programming Languages}

Previous semantic parsing work, including commercial assistants, typically translates natural language into an intermediate representation that matches the semantics of the sentences closely~\cite{steedman2011combinatory, dblp:journals/corr/liang13, banarescu2013abstract,kollar2018alexa, perera2018multi}.  
For example, the Alexa Meaning Representation Language~\cite{kollar2018alexa, perera2018multi} is associated with a closed ontology of 20 domains, each manually tuned for accuracy. %\TODO{Is phenomena the right word}
Semantically equivalent sentences have different representations, requiring complex and expensive manual annotation by experts, who must know the details of the formalism and associated ontology.
The ontology also limits the scope of the available commands, as every parameter must be an entity in the ontology (a person, a location, etc.) and cannot be free-form text.

Our approach is to represent the capability of the virtual assistant fully and formally as a VAPL; we use a deep-learning semantic parser to translate natural language into VAPL code, which can directly be executed by the assistant. Thus, the assistant's full capability is exposed to the neural network, eliminating the need and inefficiency of an intermediate representation. The VAPL code can also be converted back into a canonical natural language sentence to confirm the program before execution. Furthermore, new capabilities can be supported by extending the VAPL.  

The ThingTalk language designed for the open-source Almond virtual assistant is an example of a VAPL~\cite{almondwww17}.  ThingTalk has one construct which has three clauses: when some event happens, get some data, and perform some action, each of which can be predicated. This construct combines primitives from the extensible runtime skill library, Thingpedia, currently consisting of over 250 APIs to Internet services and IoT devices. Despite its lean syntax, ThingTalk is expressive. It is a superset of what can be expressed with IFTTT, which has crowdsourced more than 250,000 unique compound commands~\cite{ur2016trigger}. Fig.~\ref{fig:running_example} shows how a natural-language sentence can be translated into a ThingTalk program, using the services in Thingpedia.
%There is no existing corpus of ThingTalk programs to train with, and manual annotation of a large corpus would be too expensive. 

%Even taking IFTTT as source, it is very time consuming to annotate, because the programs must be checked individually and then mapped from IFTTT APIs to Thingpedia APIs. Only a subset of IFTTT APIs are available in Thingpedia, making annotation a very lossy process.

However, the original ThingTalk was not amenable to natural language translation, and no usable semantic parser has been developed.  In attempting to create an effective semantic parser for ThingTalk, we discovered important design principles for VAPL, such as matching the non-developers' mental model and keeping the semantics of components orthogonal. Also, VAPL programs must have a (unique) canonical form so the result of the neural network can be checked for correctness easily.  
We applied these principles to overhaul and extend the design of ThingTalk. Unless noted otherwise, we use ThingTalk to refer to the new design in the rest of the paper.

\subsection{Training Data Acquisition}

%The power, complexity and accuracy of semantic parsers depend both on the size, variety and correctness of the training set, and on the target language of choice.

\begin{figure*}[tb]
\includegraphics[width=\linewidth]{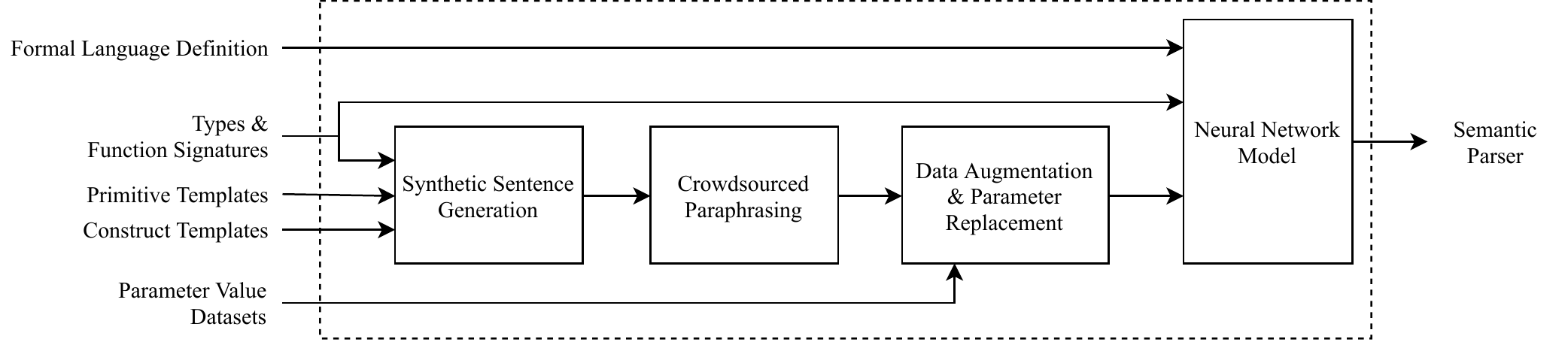}
\caption{Overview of the Genie Semantic Parser Generator}
\label{fig:overview}
\vspace{-0.5em}
\end{figure*}

%Machine learning requires training data, which is not available for new languages. Manually annotated data can be acquired in small amounts for development and testing.  How do we acquire a high-quality training corpus at a reasonable cost? 

%, how do we acquire high-quality data to create a large training dataset, we must resort to some form of automatic data generation.

%At the other end of the spectrum, previous academic work has attempted to use existing web data, such as by scraping the If-This-Then-That (IFTTT) service~\cite{ifttt,quirk15}. Through this process, they have collected a moderately large dataset of 114,408 pairs of sentence and program. Yet, they have found the data so noisy to be impossible to test on; to create a test set, they have resorted to manual annotation through crowdsourcing, and found that only 17\% of the annotated test data could be usable.

Virtual assistant development is labor-intensive, with Alexa boasting a workforce of 10,000 employees~\cite{alexa10k}.  Obtaining training data for the semantic parser is one of the challenging tasks.  How do we get training data before deployment? How can we reduce the cost of annotating usage data?  Wang et al.~\cite{overnight} propose a solution to acquire training data for the task of question answering over simple domains.  They
use a syntax-driven approach to create a canonical sentence for each formal program, ask crowdsourced workers to paraphrase canonical sentences to make them more natural, then use the paraphrases to train a machine learning model that can match input sentences against possible canonical sentences. Wang et al.'s approach designs each domain ontology individually, and each domain is small enough that all possible logical forms can be enumerated up to a certain depth. 

This approach was used in the original ThingTalk semantic parser and has been shown to be inadequate~\cite{almondwww17}.  It is infeasible to collect paraphrases for all the sentences supported by a \NLPL language.  
Virtual assistants have powerful constructs to connect many diverse domains, and their capability scales superlinearly with the addition of APIs. 
Even with our small Thingpedia, ThingTalk supports hundreds of thousands of distinct programs. 
Also, it is not possible to generate just one canonical natural language that can be understood across different domains. Crowdworkers often paraphrase sentences incorrectly or just make minor modifications to original sentences.

Our approach is to design a {\em NL-template language} to help developers data-engineer a good training set.  This language lets developers capture common ways in which VAPL programs are expressed in natural language. The NL-templates are used to {\em synthesize} pairs of natural language sentences and their corresponding VAPL code.  A sample of such sentences is paraphrased by crowdsource workers to make them more natural.  The paraphrases further inform more useful templates, which in turn derives more diverse sentences for paraphrasing.  This iterative process increases the cost-effectiveness of paraphrasing. 

Whereas the traditional approach is only to train with paraphrase data, we are the first to add {\em synthesized} sentences to the training set.  It is infeasible to exhaustively paraphrase all possible VAPL programs.  The large set of synthesized, albeit clunky, natural language commands are useful to teach the neural model compositionality. 

\iffalse
\subsection{The Genie Toolkit}
We encapsulate the semantic parser methodology we developed in a toolkit called Genie. 
Given a definition of a \NLPL language and a set of NL-templates, Genie generates a semantic parser that translates natural language to code for the new language. As shown in Figure~\ref{fig:overview}, it creates a synthesized data set, crowdsources paraphrases, and augments the training data with parameter values. It trains a neural network model and outputs the parameters of the trained model, and in other words, the semantic parser. 
% The semantic parser is then evaluated on the development set, at which point the developer can iterate until satisfied, by adding new NL-templates and crowdsourcing paraphrases again.
\fi

\subsection{Contributions}

The results of this paper include the following contributions: 

\begin{enumerate}
\item
We present the design principles for \NLPL{}s that improve the success of semantic parsing.
We have applied those principles to ThingTalk, and created a semantic parser that achieves a 62\% accuracy on data that reflects realistic Almond usage, and 63\% on manually annotated IFTTT data.
Our work is the first semantic parser for a VAPL that is extensible and supports free-form text parameters.

\item
A novel NL-template language that lets developers direct the synthesis of training data for semantic parsers of \NLPL languages.

\item
The first toolkit, Genie, that can generate semantic parsers for \NLPL languages with the help of crowdsourced workers.
%It is the first semantic parser that accepts natural language to formal language translation rules to improve the quality of synthetic sentence generation.
As shown in Fig.~\ref{fig:overview}, Genie accepts a VAPL language and a set of NL-templates. Genie
adds synthesized data to paraphrased data in training, expands parameters, and pretrains a language model.  
Our neural model achieves an improvement in accuracy of 17\% on IFTTT and 15\% on realistic inputs, compared to training with paraphrase data alone.
% 16% is correct on overall-test (which combines IFTTT test and Cheatsheet test); paraonly overall is 47.9, and combined 62.5; the difference is 15.6 which rounds to 16
% this is probably not obvious from the text though, so let's show different numbers for IFTTT and Cheatsheet

%%introduces the use of max-margin and span-level copying mechanism, which each accounts for a 3\% improvement in accuracy.  
\item
Demonstration of extensibility by using Genie to generate effective semantic parsers for a Spotify skill, access control, and aggregate functions.
\end{enumerate}

\subsection{Paper Organization}
The organization of the paper is as follows. Section~\ref{sec:vapl-principles} describes the \NLPL principles, using \TT as an example.
Section~\ref{sec:dataset} introduces the Genie data acquisition pipeline, and Section~\ref{sec:nl-model} describe the semantic parsing model used by Genie.
We present experimental results on understanding \TT commands in Section~\ref{sec:eval},
and additional languages and case studies in Section~\ref{sec:case-studies}.
We present related work in Section~\ref{sec:related} and conclusion in Section~\ref{sec:conclusion2}.

%\input{problem}
%!TEX root = ./paper.tex
\section{Principles of \NLPL Design}
\label{sec:vapl-principles}

Here we discuss the design principles of Virtual Assistant Programming Languages (\NLPL) to make it amenable to natural language translation, using ThinkTalk as an example. 
%\TT is also the base language on which new \NLPL languages can be defined using Genie.
These design principles can be summarized as (1) strong fine-grained typing to guide parsing and dataset generation,
(2) a skill library with semantically orthogonal components designed to reduce ambiguity, (3) matching the user's mental model to simplify the translation, and (4)
canonicalization of \NLPL programs.

\subsection{Strong, Static Typing}
To improve the compositionality of the semantic parser, \NLPL{}s should be statically typed. They should include
fine-grain domain-specific types and support definitions of custom types.
The ThinkTalk type system includes the standard types: strings, numbers, booleans, and enumerated types.
It also has native support for common object types used in IoT devices and web services.
%such as measurements, locations, pictures, currencies, hashtags, usernames, path names, URLs, dates, times, phone numbers and email addresses.
Developers can also provide custom \textit{entity types}, which are internally represented as opaque identifiers but can be
recalled by name in natural language. 
Arrays are the only compound type supported.

To allow translation from natural language without contextual information, the \NLPL also needs a rich
language for constants.
\iffalse
Locations can be represented as absolute points on Earth, or as the special
constants $\texttt{here}$, $\texttt{home}$ and $\texttt{work}$, which represent the user's current
location, home address and work address, respectively. Dates and times can be specified absolutely,
or as an offset from an absolute date-time or the current time (eg. ``3 hours ago'' is represented as $\texttt{now} - 3\text{h}$),
or as the beginning or end of a time unit (eg. ``today'' is translated to $\texttt{start\_of\_day}$).
\fi
For example, in ThingTalk, measures can be represented with any legal unit, and can be composed additively (as in ``6 feet 3 inches'', which is
translated to $6\text{ft} + 3\text{in}$); this is necessary because a neural semantic parser cannot
perform arithmetic to normalize the unit during the translation.

Arguments such as numbers, dates and times, in the input sentence are identified and normalized using a rule-based algorithm~\cite{dong2016language}; they are replaced as named constants of the form ``NUMBER\_0'', ``DATE\_1'', etc.
String and named entity parameters instead are represented
using multiple tokens, one for each word in the string or entity name; this allows the words to be copied from the input
sentence individually. Named entities are normalized with a knowledge base lookup after parsing.

%Numbers, dates and times, which can be identified and normalized using a rule-based algorithm in the input sentence,
%are represented as named constants of the form ``NUMBER\_0'', ``DATE\_1'', etc., using a technique called
%\textit{argument identification}~\cite{dong2016language}. String and named entity parameters instead are represented
%using multiple tokens, one for each word in the string or entity name; this allows the words to be copied from the input
%sentence individually. Named entities are normalized with a knowledge base lookup after parsing.

\subsection{Skill Library Design}

The skill library defines the virtual assistant's knowledge of the Internet services and IoTs. As such, the design of the representation as well as how information is organized are very important. In the following, we describe a high-level change we made to the original ThingTalk, then present the new syntax of classes and the design rationale. 

\paragraph{Orthogonality in Function Types.}
%When we ported over the Thingpedia library from ThingTalk to the \TT, we made a major change to the representation.  
In the original Thingpedia library, there are three kinds of functions: 
{\em triggers}, {\em retrievals}, and {\em actions}~\cite{almondwww17}.  Triggers are callbacks or polling functions that return results upon the arrival of some event, retrievals return results, actions create side effects.  Unfortunately, the semantics of callbacks and queries are not easily discernible by consumers.  It is hard for users to understand that ``when Trump tweets'' and ``get a Trump tweet'' refer to two different functions. Furthermore, when a device supports only a trigger and not a retrieval, or vice versa, it is not apparent to the user which is allowed.  The inconsistency in functionality makes getting correct training and evaluation data problematic.

In the new ThingTalk, we collapse the distinction between triggers and retrievals into one class: queries, which can be monitored as an event or used to get data.  The runtime ensures that any supported retrieval can be monitored as events, and vice versa. Not only does this provide more functionality, it also makes the language more regular and hence simpler for the users, crowdsourced paraphrase providers, and the neural model.  

\begin{figure}
\fontsize{8}{10}\selectfont
\begin{tabbing}
\=123456789012345678901234\=\kill
\>Class $c$: \> $\texttt{class}~~@\textit{cn}~~\left[\texttt{extends}~~@\textit{cn}\right]^*~~\{~~\left[\textit{qd}\right]^*\left[\textit{ad}\right]^*\}$\\
\>Class name $\textit{cn}$: \>identifier\\
\>Query declaration $\textit{qd}$: \> $\texttt{monitorable}?~~\texttt{list}?~~\texttt{query}~~\textit{fn}( \left[\textit{pd}\right]^*);$\\
\>Action declaration $\textit{ad}$: \> $\texttt{action}~~\textit{fn}( \left[\textit{pd}\right]^*);$\\
\>Function name $\textit{fn}$: \>identifier\\
\>Parameter declaration $\textit{pd}$: \> $\left[\texttt{in req}~\vert~\texttt{in opt}~\vert~\texttt{out}\right]~~\textit{pn}:\textit{t}$\\
\>Parameter name $\textit{pn}$: \>identifier\\
\>Parameter type $t$: \> $\texttt{String}~~\vert~~\texttt{Number}~~\vert~~\texttt{Boolean}~~\vert~~\texttt{Enum}([v]^+)~~\vert$\\
        \>\> $\texttt{Measure}(u)~~\vert~~\texttt{Date}~~\vert~~\texttt{PathName}~~\vert$\\
        \>\> $\texttt{Entity}(\textit{et})~~\vert~~\texttt{Array}(t)~~\vert~~\ldots$\\
\>Value $v$: \> $\text{literal}$\\
\>Unit $u$: \> $\text{bytes}~~\vert~~\text{KB}~~\vert~~\text{ms}~~\vert~~\text{s}~~\vert~~\text{m}~~\vert~~\text{km}~~\vert~~\ldots$\\
\>Entity type $\textit{et}$: \> $\text{literal}$
\end{tabbing}
\vspace{-1em}
\caption{The formal grammar of classes in the skill library.}
\label{fig:class-grammar}
\vspace{-1em}
\end{figure}

\paragraph{Skill Definition Syntax.}
For modularity, every \NLPL should include a skill library, a set of classes representing the domains and operations
supported. In \TT, skills are IoT devices or web services. 
The formal grammar of \TT classes is shown in Fig.~\ref{fig:class-grammar}.

Classes have functions, belonging to one of two kinds: \textit{query functions} retrieve data and have no side-effects; 
\textit{action functions} have side-effects but do not return data.
% The distinction of query functions and action functions is important for training data synthesis and optimizations \TODO{don't know what optimizations refer to}. 
% Queries can be rearranged and modified without visible side effects, provided that the actions are called the current
% number of times and in the right order.

A query function can be \textit{monitorable}, which means that the result returned can be monitored for changes.  The result can be polled, or the query function supports push notifications.  Queries that cannot be monitored include those that change constantly, such as the retrieval of a random cat picture used in Fig.~\ref{fig:running_example}.  
A query can return a single result or return a \textit{list} of results.
%Each parameter indicates its type statically. 

% Genie only needs the signatures of the functions; their implementations are used by the run-time of the \NLPL languages and are not needed for semantic parsing.
The function signature includes the class name, the function name, the type of the function, all the parameters and their types.  
Data are passed in and out of the functions through named parameters, which can be required or optional.
Action methods have only input parameters, while query methods have both input and output parameters.

% Functions are also annotated with natural language: a canonical name for the function and each parameter.

An example of a class, for the Dropbox service, is shown in
Fig.~\ref{fig:dropbox-class}.
It defines three queries: ``get\_space\_usage'', ``list\_folder'', and ``open'', and an action ``move''.
The first two queries are monitorable; the third returns a randomized download link for each invocation
so it is not monitorable.

\begin{figure}
\fontsize{8}{10}\selectfont
\begin{tabbing}
\=12345678901\=2\=34567890123456789012\=34\=\kill
\textbf{class} @com.dropbox \{\\
\>$\textbf{monitorable query}~\text{get\_space\_usage}(\texttt{out}~~\textit{used\_space}: \text{Measure}(\text{byte}),$\\
\>\>\>\>\>$\texttt{out}~~\textit{total\_space}: \text{Measure}(\text{byte}));$\\
\>$\textbf{monitorable list query}~\text{list\_folder}(\texttt{in req}~~\textit{folder\_name}: \text{PathName},$\\
\>\>\>\>$\texttt{in opt}~~\textit{order\_by}: \text{Enum},$\\
\>\>\>\>$\texttt{out}~~\textit{file\_name}: \text{PathName},$\\
\>\>\>\>$\texttt{out}~~\textit{is\_folder}: \text{Boolean},$\\
\>\>\>\>$\texttt{out}~~\textit{modified\_time}: \text{Date},$\\
\>\>\>\>$\texttt{out}~~\textit{file\_size}: \text{Measure}(\text{byte}),$\\
\>\>\>\>$\texttt{out}~~\textit{full\_path}: \text{PathName});$\\
\>$\textbf{query}~\text{open}(\texttt{in req}~~\textit{file\_name}: \text{PathName},$\\
\>\>$\texttt{out}~~\textit{download\_url}: \text{URL});$\\
\>$\textbf{action}~\text{move}(\texttt{in req}~~\textit{old\_name}: \text{PathName},$\\
\>\>\>$\texttt{in req}~~\textit{new\_name}: \text{PathName});$\\
\>$\ldots~~\}$\\
\end{tabbing}
\vspace{-2em}
\caption{The Dropbox class in the \TT skill library.}
\label{fig:dropbox-class}
\end{figure}

\subsection{Constructs Matching the User's Mental Model}

\NLPL{}s should be designed to reflect how users typically specify commands.
It is more natural for users to think about data with certain characteristics, rather than execution paths.
For this reason, ThingTalk is data focused and not control-flow focused.
%\TT is
ThingTalk has a single construct:
\begin{small}
\begin{align}
s\Rightarrow q\Rightarrow a\nonumber
\end{align}
\end{small}The \textit{stream} clause, $s$, specifies the evaluation of the program as a continuous stream
of events. The optional \textit{query} clause, $q$, specifies what data should be retrieved when the events occur.
The \textit{action} clause, $a$, specifies what the program should do. 
The formal grammar is shown in Fig.~\ref{fig:tt-grammar}.

\begin{figure}
\fontsize{8}{10}\selectfont
\begin{tabbing}
12\=123456789012345678901\=\kill
\>Program $\pi$: \> $s \left[\Rightarrow q \right]? \Rightarrow a;$\\
\>Stream $s$: \>$\texttt{now}~~\vert~~\texttt{attimer}~~\texttt{time} = v~~\vert$\\
        \>\>$\texttt{timer}~~\texttt{base} = v~~\texttt{interval} = v~~\vert$\\
        \>\>$\texttt{monitor}~q~~\vert~~\texttt{edge}~~s~~\texttt{on}~~p$\\
\>Query $q$: \>$f~~\left[\textit{ip} = v\right]^*~~\left[\textit{ip} = \textit{op}\right]^*~~\vert~~q~~\texttt{filter}~~p~~\vert$\\
        \>\>$q~~\texttt{join}~~q~~\left[\texttt{on}~~\left[\textit{ip} = \textit{op}\right]^+\right]?$\\
\>Action $a$: \>$f~~\left[\textit{ip} = v\right]^*~~\left[\textit{ip} = \textit{op}\right]^*~\vert~~\texttt{notify}$\\
\>Function $f$: \>$@\textit{cn}.\textit{fn}$\\
\>Class name $\textit{cn}$: \>identifier\\
\>Function name $\textit{fn}$: \>identifier\\
\>Input parameter $\textit{ip}$: \> $\textit{pn}:t$\\
\>Output parameter $\textit{op}$: \> $\textit{pn}:t$\\
\>Parameter name $\textit{pn}$: \>identifier\\
\>Parameter type $t$: \> the parameter type in Fig.~\ref{fig:class-grammar}\\
\>Predicate $p$: \>$\texttt{true}~~\vert~~\texttt{false}~~\vert~~\texttt{!}\textit{p}~~\vert~~\textit{p}\texttt{ \&\& }\textit{p}~~\vert~~
\textit{p}~\texttt{||}~\textit{p}~~\vert$\\
        \>\>$\textit{op}~\textit{operator}~v~~\vert~~f~~\left[\textit{ip} = v\right]^*~\texttt{\{}~~p~~\texttt{\}}$\\
\>Value $v$: \>$\text{literal}~~\vert~~\texttt{enum}:\text{identifier}$\\
\>Operator $\textit{operator}$: \> $\text{==}~~\vert~~\text{>}~~\vert~~\text{<}~~\vert~~
\texttt{contains}~~\vert~~\texttt{substr}~~\vert$\\
        \>\>$\texttt{starts\_with}~~\vert~~\texttt{ends\_with}~~\vert~~\ldots$
\end{tabbing}
\vspace{-1em}
\caption{The formal grammar of \TT.}
\label{fig:tt-grammar}
\vspace{-1em}
\end{figure}

%\TODO{shouldn't we use s => q => a, and then say ``The stream clause, s, specifies the .... ``.  This prepares users with the notation s=>q=>a later.}
% GC: it is less readable this way, but I agree it makes the notation consistent with later

An example illustrating parameter passing is 
shown in Fig.~\ref{fig:running_example}.  The use of filters is demonstrated with 
the following example that automatically retweets the tweets from PLDI:

{\small
\begin{tabbing}
123\=12345678\=\kill
\>$\texttt{monitor}(@\text{com.twitter.timeline}()\texttt{ filter }\textit{author}=\text{@PLDI})$\\
$\Rightarrow$\>$@\text{com.twitter.retweet}(\textit{tweet\_id}=\textit{tweet\_id})$
\end{tabbing}}
The program uses the $\textit{author}$ output parameter of the function $@\text{com.twitter.timeline}()$ to
filter the set of tweets, and passes the $\textit{tweet\_id}$ output parameter to the input parameter with the same
name of $@\text{com.twitter.retweet}()$.

%\TODO{Show a couple of examples, with parameters, filters, mention parameters, filters.  I don't think the readers can infer examples of applications from the grammar, and appreciate all the discussions below}

\paragraph{Queries and Actions.}

To match the user's mental model, ThingTalk uses implicit, rather than explicit, looping constructs. The user is given the abstraction that they are describing operations on scalars.  In reality, queries always return a list of results; functions returning a single result are converted to return singleton lists. These lists are implicitly traversed; each result can be used as an input parameter in a subsequent function invocation.

The result of queries can be optionally filtered with a boolean predicate using
equality, comparions, string and array containment operators, as well as predicated query functions.
For example, the following retrieves only ``emails from Alice'':
{\small
\begin{tabbing}
1234\=567\=12345678\=\kill
$\texttt{now}$\>$\Rightarrow$\>$(@\text{com.gmail.inbox}())~~\texttt{filter}~~\textit{sender}=\text{``Alice''}\Rightarrow\texttt{notify}$
\end{tabbing}}

We introduce the \textit{join} operator for queries in ThingTalk to support multiple retrievals in a program. 
When joining two queries, parameters can be passed between the queries, and the results are the cross product of the respective queries. A program's action can be either the builtin \texttt{notify}, which presents the result to the user, or an action function defined in the library.

For example, the following ``translates the title of New York Times articles'':
{\small
\begin{tabbing}
1234\=567\=12345678\=\kill
$\texttt{now}$\>$\Rightarrow$\>$@\text{com.nytimes.get\_front\_page}()~~\texttt{join }$\\
\>\>$@\text{com.yandex.translate}()\texttt{ on }\textit{text}=\textit{title}\Rightarrow\texttt{notify}$
\end{tabbing}}

%or search ``get the ArXiv link of papers mentioned in my Twitter feed''.

\iffalse
Action functions have only input parameters, while query functions have both input and output parameters.
Output parameters can be passed from one function to the next.
\fi

\paragraph{Streams.}
Streams are a new concept we introduce to ThingTalk. They generalize the trigger concept in the original language to enable
reacting to arbitrary changes in the data accessible by the virtual assistant.
A stream can be (1) the degenerate stream ``\texttt{now}'', which triggers the program once immediately, (2) a timer, or (3) a \texttt{monitor} of a query, which triggers whenever the query result changes.
Any query that uses monitorable functions can be monitored, including queries that use joins or filters.

We introduce a stream operator, \textit{edge filter}, to watch for changes in a stream. It triggers whenever a boolean predicate on the values monitored transition from false to true; the predicate is assumed to be previously false for the first value in a stream. The program below notifies the users each time the temperature drops below the $60$ degrees Fahrenheit theshold.
\begin{small}
\begin{align}
&\texttt{ edge}\left(\texttt{monitor } @\text{weather.current}()\right)\texttt{ on }\textit{temperature}<60F\nonumber\\
\Rightarrow&\texttt{ notify};\nonumber
\end{align}
\end{small}The edge filter operator allows us to convert all previous trigger functions to the new language without losing functionality. 

\paragraph{Input and Output Parameters.}
To aid translation from natural language, ThingTalk uses keyword parameters  rather than the more conventional positional parameters.
With keyword parameters, the semantic parser needs to learn just the partial signature of the functions, and not even
the length of the signature. We annotate each parameter with its type, with the goal to help the model distinguish between parameters with the same name and to unify parameters by type.
For readability, type annotations are omitted from the examples in this paper.
%\TODO{EXPLAIN THIS BETTTER}; hence, the language model of parameters can generalize across functions.
To increase compositionality, we encourage developers to use the same naming conventions so the same parameter names are used for similar purposes. 

When an output of a function is passed into another as input, the former parameter name is simply assigned to the latter.  For example, in Fig.~\ref{fig:running_example}, the output parameter \textit{picture\_url} of $@\text{com.thecatapi.get}()$ is assigned to the input parameter of 
$@\text{com.facebook.post\_picture}()$ with the same name. 
%\TODO{Describe the cat example.}
This design avoids introducing new variables in the program so the semantic parser only needs to
learn each function's parameter names. 
If output parameters in two functions in a program have the same name, we assume that the name refers to the rightmost instance. 
Here we consciously trade-off completeness for accuracy in the common case.

\subsection{Canonicalization of Programs}
As described in Section~\ref{sec:intro}, canonicalization is key to training a neural semantic parser.  We use semantic-preserving transformation rules to give \TT programs a canonical form. 
For example, query joins without parameter passing are a commutative operation, and are canonicalized by ordering the operands lexically.
Nested applications of the \texttt{filter} operator are canonicalized to a single filter with the \texttt{\&\&} connective.
Boolean predicates are simplified to eliminate redundant expressions, converted to \textit{conjunctive normal form} and then canonicalized
by sorting the parameters and operators. Each clause is also automatically moved to the left-most function that includes all the output parameters.
Input parameters are listed in alphabetical order, which helps the neural model learn a global order that is the same across all functions.

%\subsection{Design Rationale}

%\TT is designed to be elegant for users to express trigger-action commands in natural language. Natural language translation is made possible by the use of very high-level functions, which are annotated with a natural language description, and by the use of named parameters, which can appear in any order, and whose natural language paraphrases can be learned from data.

%\input{methodology}
%\input{thingpedia}
%\input{thingtalk}
%!TEX root = ./paper.tex
\section{Genie Data Acquisition System}
\label{sec:dataset}

The success of machine learning depends on a high-quality training set; it must represent real, correctly labeled, inputs. To address the difficulty in getting a training set for a new language, Genie gives developers (1) a novel language-based tool to synthesize data, (2) a crowdsourcing framework to collect paraphrases, (3) a large corpus of values for parameters in programs, and (4) a training strategy that combines synthesized and paraphrase data.
\subsection{Data Synthesis}

%Previous research proposed using the formal grammar to generate possible programs, and semantic rules to translate the program into a canonical NL sentence, which is then shown to crowdsource workers for paraphrases~\cite{overnight}.  We found that this does not work for virtual assistant commands. -- Already discussed.

As a programming language, ThingTalk may seem to have a small number of components: queries, streams, actions, filters, and parameters.  However, it has a library of skills belonging to many domains, each using different terminology.  
Consider the function 
``list\_folder'' in Dropbox, which returns a modified time.  If we want to ask for ``my Dropbox files that changed this week'', we can add the filter 
$\textit{modified\_time} > \texttt{start\_of\_week}$.  An automatically generated sentence would read: ``my Dropbox files having modified time after a week ago''.  Such a sentence is hard for crowdsource workers to understand.  If they do not understand it, they cannot paraphrase it. 

Similarly, even though ThingTalk has only one construct, there are various ways of expressing it.  
Here are two common ways to describe event-driven operations: ``when it rains, remind me to bring an umbrella'', or ``remind me to bring an umbrella when it rains''. Here are two ways to compose functions: ``set my profile at Twitter with my profile at Facebook'' or ``get my profile from Facebook and use that as a profile at Twitter''.  

We have created a {\em NL-template language} so developers can generate variety when synthesizing data.  
We hypothesize that we can exploit compositionality in natural language to factor data synthesis into primitive templates for skills and construct templates for the language. From 
a few templates per skill and a few templates per construct component, we hope to generate a representative set of synthesized sentences that are understandable. 
\iffalse
\begin{figure}
\centering
\includegraphics[width=0.9\linewidth]{figures/data-collection.pdf}
\caption{Training set data generation}
\label{fig:gen_synthetic}
\end{figure}
\fi

\begin{table}
\fontsize{8}{10}\selectfont
\centering
{\setlength\tabcolsep{2pt}
\begin{tabular}{p{2.8cm}|c|p{4.6cm}}
\toprule
\bf Natural language & \bf Cat. & \bf \TT Code\\
\hline
my Dropbox files & \NOUNPHRASE & $@\text{com.dropbox.list\_folder}()$\\
%\hline
%my Dropbox files in alphabetical order  & \NOUNPHRASE & $@\text{com.dropbox.list\_folder}(\textit{order\_by}$ $=\text{name\_increasing})$\\
\hline
my Dropbox files that changed most recently  & \NOUNPHRASE & $@\text{com.dropbox.list\_folder}(\textit{order\_by}$ $=\text{modified\_time\_decreasing})$ \\
\hline 
my Dropbox files that changed this week & \NOUNPHRASE & $@\text{com.dropbox.list\_folder}(\textit{order\_by}=\text{modified\_time\_decreasing}) \texttt{ filter}$ $\textit{modified\_time} > \texttt{start\_of\_week}$ \\
%\hline 
%my Dropbox files larger than \$x & \NOUNPHRASE & $\lambda(x : \text{Measure}(\text{byte})) \rightarrow \newline @\text{com.dropbox.list\_folder}()$ $\texttt{filter }\textit{file\_size} > x$ \\
\hline
files in my Dropbox folder \$x & \NOUNPHRASE & $\lambda(x : \text{PathName}) \rightarrow \newline @\text{com.dropbox.list\_folder}(\textit{folder\_name}$ $=x)$ \\
%\hline
%\mbox{subfolders of \$folder} \mbox{ in my Dropbox} & $@\text{com.dropbox.list\_folder}(\textit{folder\_name}$ \mbox{$=\text{\$folder}) \texttt{ filter }\textit{is\_folder}=\texttt{true}$}\\
\hline
\mbox{when I modify a file} in Dropbox & \WHPHRASE & $\texttt{monitor }@\text{com.dropbox.list\_folder}()$ \\
\hline
when I create a file in Dropbox & \WHPHRASE & $\texttt{monitor }@\text{com.dropbox.list\_folder}()$ $\texttt{on new }\textit{file\_name}$\\
\hline
the download URL of \$x & \NOUNPHRASE & \multirow{4}{5cm}{$\lambda(x : \text{PathName}) \rightarrow \newline @\text{com.dropbox.open}(\textit{file\_name}$ $=x)$}\\
\cline{1-2}
a temporary link to \$x & \NOUNPHRASE \\%$\lambda(\text{p\_file\_name} : \text{PathName}) \rightarrow \newline @\text{com.dropbox.open}(\textit{file\_name}$ $=\text{p\_file\_name})$\\
\cline{1-2}
open \$x & \VERBPHRASE \\%$\lambda(\text{p\_file\_name} : \text{PathName}) \rightarrow \newline @\text{com.dropbox.open}(\textit{file\_name}$ $=\text{p\_file\_name})$\\
\cline{1-2}
download \$x & \VERBPHRASE \\%& $\lambda(\text{p\_file\_name} : \text{PathName}) \rightarrow \newline @\text{com.dropbox.open}(\textit{file\_name}$ $=\text{p\_file\_name})$\\
\bottomrule
\end{tabular}
}
\caption{Examples of developer-supplied primitive templates for the @com.dropbox.list\_folder and @com.dropbox.open functions in the Thingpedia library, with their grammar category.
\NOUNPHRASE, \WHPHRASE, and \VERBPHRASE refer to \textit{noun phrase}, \textit{when phrase} and \textit{verb phrase}, respectively.}
\label{table:dropbox}
\vspace{-2em}
\end{table}

\iffalse
\begin{figure}
\fontsize{8}{10}\selectfont
\begin{flalign*}
&\text{please } \text{\VERBPHRASE}[a] \rightarrow \text{\COMMAND}[\texttt{now} \Rightarrow a]\\
&\text{show me } \text{\NOUNPHRASE}[q]  \rightarrow \text{\COMMAND}[\texttt{now} \Rightarrow q \Rightarrow \texttt{notify}]\\
&\text{list } \text{\NOUNPHRASE}[q]  \rightarrow \text{\COMMAND}[\texttt{now} \Rightarrow q \Rightarrow \texttt{notify}] \text{ if }q\text{ is a list}\\
&\text{monitor } \text{\NOUNPHRASE}[q] \rightarrow \text{\COMMAND}[\texttt{monitor}~~ q \Rightarrow \texttt{notify}]\\
&\text{get } \text{\NOUNPHRASE}[q] \text{ and then } \text{\VERBPHRASE}[a] \rightarrow \text{\COMMAND}[\texttt{now} \Rightarrow q \Rightarrow a]\\
&\text{when } \text{\NOUNPHRASE}[q] \text{ changes } \rightarrow \WHPHRASE[\texttt{monitor}~~q]\\
&\text{every } \text{\INTERVAL}[i] \rightarrow \WHPHRASE[\texttt{timer}(\textit{interval}=i)]\\
&\text{alert me } \text{\WHPHRASE}[s] \rightarrow \text{\COMMAND}[s \Rightarrow \text{notify}]\\
&\text{\WHPHRASE}[s] \text{ , } \text{\VERBPHRASE}[q] \rightarrow \text{\COMMAND}[s \Righatrrow a]\\
&\text{\VERBPHRASE}[q] \text{ }\text{\WHPHRASE}[s] \rightarrow \text{\COMMAND}[s \Rightarrow a]\\
\end{flalign*}
\vspace{-1.5em}
\caption{The template grammar used by the semantic parser. In square brackets on the right hand-side at the semantics of the generated derivation.
VP, NP and WP refer to \textit{verb phrase}, \textit{noun phrase} and \textit{when phrase} respectively. 
\label{fig:generic-grammar}
\end{figure}
\fi

\paragraph{Primitive Templates.}

Genie allows the skill developer to provide a list of \textit{primitive templates}, each of which consists of code using that skill, a natural language utterance describing it, and the natural language grammar category of the utterance. 
The templates define how the function should be invoked, and how parameters are passed and used.
The syntax is as follows:
\begin{small}
\begin{align*}
\textit{cat} \texttt{ := } u \rightarrow \lambda([\textit{pn}:t]^*) \rightarrow [s~~|~~q~~|~~a]
\end{align*}%Note a line break after small add a space before the word This
\end{small}This syntax declares that the utterance $u$, belonging to the grammar category $\textit{cat}$ (\textit{verb phrase}, \textit{noun phrase}, or \textit{when phrase}), maps to stream $s$, query $q$ or action $a$. 
The utterance may include placeholders, prefixed with \$, which are used as parameters, $\textit{pn}$ of type $t$,  
in the stream, query or action. 

Examples of primitive templates for functions $@\text{com.dropbox.list\_folder}$ and $@\text{com.dropbox.open}$
are shown in Table~\ref{table:dropbox}.  Multiple templates can be provided for the same function, as different combinations
of input and output parameters can provide different semantics, and functions can be filtered and monitored as well.

Note that the function $@\text{com.dropbox.open}$ is a query, not an action, because it returns a result (the download link);
the example shows that utterances for queries can be both noun phrases (``the download URL'') and verb phrases (``open'').
The ability to map the same program fragment to different grammar categories is new in Genie, and differs from the knowledge base representation previously used by Wang et al.~\cite{overnight}.
Other examples of verb phrases for queries are ``translate \$x'' (same as ``the translation of \$x'') and ``describe \$x'' (same as ``the description of \$x'').

While designing primitive templates, it is important to choose grammar categories that compose naturally.
The original design of ThingTalk exclusively used verb phrases for queries; we switched to mostly noun phrases as they can be substituted as input parameters (e.g., ``a cat picture'' can be substituted for parameter $\$x$ in ``post $\$x$ on Twitter'' and ``post $\$x$ on Facebook'').

\paragraph{Construct Templates.}

To combine the primitives into full programs, the language designer also
provides a set of \textit{construct templates}, mapping natural language compositional constructs to 
formal language operators. 
A construct template has the form:
\begin{small}
\begin{align*}
\textit{lhs}\texttt{ := }\left[\text{literal}~~\vert~~\textit{vn} : \textit{rhs}\right]^+ \rightarrow \textit{sf}
\end{align*}
\end{small}which says that a derivation of non-terminal category \textit{lhs} can be constructed by
combining the literals and variables \textit{vn} of non-terminal category \textit{rhs}, and then applying the semantic
function \textit{sf} to compute the formal language representation.
For example, the following two construct templates define the
two common ways to express ``when - do'' commands:

{\small
\begin{tabbing}
1234\=\kill
\>$\COMMAND := s : \text{\WHPHRASE} \text{ `,' } a: \text{\VERBPHRASE} \rightarrow \textbf{return } s \Rightarrow a;$\\
\>$\COMMAND := a: \text{\VERBPHRASE} ~~ s : \text{\WHPHRASE} \rightarrow \textbf{return } s \Rightarrow a;$
\end{tabbing}}%
\noindent Together with the following two primitive templates:

{\small
\begin{tabbing}
1234\=123\=123\=\kill
\>$\WHPHRASE$\>$:= \text{`when I modify a file in Dropbox'} \rightarrow$\\
\>\>\>$\texttt{monitor }@\text{com.dropbox.list\_folder}()$\\
\>$\VERBPHRASE$\>$:= \text{`send a Slack message'} \rightarrow @\text{com.slack.send}()$
\end{tabbing}
}%
\noindent Genie would generate the commands ``when I modify a file in Dropbox, send a Slack message'' and
``send a Slack message when I modify a file in Dropbox''.

Semantic functions allow developers to write arbitrary code that computes the formal
representation of the generated natural language.  The following performs type-checking, ensuring that only monitorable queries are monitored.
%\TODO{Do we need to add edge filters?}
% GC: for what? they are just another template construct; we're not showing all constructs anyway
% using the variables intermediate formal language components named between \texttt{} and $\rightarrow$. For example, semantic functions can apply incremental type-checking:

{\small
\begin{tabbing}
1234\=1234\=1234\=\kill
\>$\WHPHRASE := \text{`when' } q: \text{\NOUNPHRASE} \text{ `change' } \rightarrow \{$\\
\>\>$\textbf{if}~~q.\text{is\_monitorable}$\\
\>\>\>$\textbf{return }\texttt{monitor}~~q$\\
\>\>$\textbf{else}$\\
\>\>\>$\textbf{return }\perp$\\
\>$\}$
\end{tabbing}}%

For another example, the following checks that the argument is a list, so as to  be compatible with the semantics of the verb ``enumerate'':

{\small
\begin{tabbing}
1234\=1234\=1234\=\kill
\>$\COMMAND := \text{`enumerate' } q: \text{\NOUNPHRASE} \rightarrow \{$\\
\>\>$\textbf{if}~~q.\text{is\_list}$\\
\>\>\>$\textbf{return }\texttt{now} \Rightarrow q \Rightarrow \texttt{notify}$\\
\>\>$\textbf{else}$\\
\>\>\>$\textbf{return }\perp$\\
\>$\}$
\end{tabbing}
}%

Each template can also optionally be annotated with a boolean flag, which allows the developer to define different subsets of rules for different purposes (such as training or paraphrasing).

\paragraph{Synthesis by Sampling}
Previous work by Wang et al.~\cite{overnight} 
recursively enumerates \textit{all} possible derivations, up to a certain depth of the derivation tree. Such an approach
is unsuitable for ThingTalk, because the number of derivations grows exponentially with increasing depth and library size. 
Instead, Genie uses a randomized synthesis algorithm, which considers only   
a subset of derivations produced by each construct template. The desired size is configurable, and the number of derivations decreases exponentially with increasing depth.  The large number of low-depth programs provide breadth, and the relatively smaller number of high-depth programs add variance to the synthesized set and expand the set of recognized programs. 
Developers using Genie can control the sampling strategy by splitting or combining construct templates, using intermediate derivations. For example, the following template:

{\small
\begin{tabbing}
1234\=123456789012\=\kill
\>$\textsc{np} := q: \text{\NOUNPHRASE} \text{ `having' } p: \textsc{pred} \rightarrow\textbf{return } q~~\texttt{filter}~~p$\\
\>$\COMMAND := \text{`get' } q: \textsc{np} \text{ `and then' } a: \text{\VERBPHRASE}$\\
\>\>$ \rightarrow\textbf{return } \texttt{now} \Rightarrow q \Rightarrow a$\\
\end{tabbing}
}
\noindent can be combined in a single template that omits the intermediate noun phrase derivation:

{\small
\begin{tabbing}
1234\=123456789012\=\kill
\>$\COMMAND :=$\> $\text{`get' } q: \textsc{np} \text{ `having' } p: \textsc{pred} \text{ `and then' } a: \text{\VERBPHRASE}$\\
\>\>$\rightarrow\textbf{return } \texttt{now} \Rightarrow q~~\texttt{filter}~~p \Rightarrow a$\\
\end{tabbing}
}(\textsc{pred} is the grammar category of boolean predicates.)
The combined template has lower depth, so more sentences would be sampled from it. Conversely, if a single template is split into two or more templates, the number of sentences using the construct becomes lower.

\subsection{Paraphrase Data}

Sentences synthesized from templates are not representative of real human input, thus we ask crowdsource workers to rephrase them in more natural sentences. However, manual paraphrasing is not only expensive, but it is also error-prone, especially when the synthesized commands are complex and do not make sense to humans. Thus, Genie lets the developer decide which 
synthesized sentences to paraphrase.  Genie also has a crowdsourcing framework designed to improve the quality of paraphrasing. 

\paragraph{Choosing Sentences to Paraphrase.}
Developers can control the subset of templates to paraphrase as well as their sampling rates. 
It is advisable to obtain some paraphrases for every primitive, but combinations of functions need not be sampled evenly, since coverage is provided by including the synthesized data in training.  
Our priority is to choose sentences that workers can understand and can provide a high-quality paraphrase.  

%On the other hand, they can understand combinations where the result of one function is passed as input to another.
Developers can provide lists of easy-to-understand and hard-to-understand functions. 
We can maximize the success of paraphrasing, while providing some coverage, by creating compound sentences that combine the easy functions with difficult ones. We avoid combining unrelated functions because they confuse workers.

Developers can also specify input parameter values to make the synthesized sentences easier to understand.  String parameters are quoted, Twitter usernames have @-signs, etc, so workers can identify them as such and copy them in the paraphrased sentences properly. (Note that quotes are removed before they are used for training).

\paragraph{Crowdsourcing.}
Genie also automates the process of crowdsourcing paraphrases. Based on the selected set of synthesized sentences to paraphrase,
Genie produces a file that can be used to create a batch of crowdsource tasks on the Amazon Mechanical Turk platform.
% Workers are shown the synthesized sentence and hints, and are asked to enter the paraphrase. 
To increase variety, Genie prepares the crowdsource tasks so that multiple workers see the same synthesized sentence, and
each worker is asked to provide two paraphrases for each sentence.  We found that people will only make the most obvious change if asked to provide one paraphrase, and they have a hard time writing three different paraphrases.

Due to ambiguity in natural language and workers not reading instructions and performing minimal work etc., the answers provided can be wrong.  Genie uses heuristics to discard obvious mistakes, and asks workers to check the correctness of remaining answers.

\subsection{Parameter Replacement \& Data Augmentation}

During training, it is important that the model sees many different combinations of parameter
values, so as not to overfit on specific values present in the training set.
Genie has a built-in database containing 49 different parameter lists and gazettes of named entities, including
corpora of YouTube video titles and channel names, Twitter
and Instagram hashtags, song titles, people names, country names, currencies, etc. These corpora were collected from various resources on the Web
and from previous academic efforts~\cite{yang2011patterns, shetty2004enron, francis1971manual, almeida2011contributions, Dua:2017, gasparetti2017modeling, leskovec2009meme, DBLP:journals/corr/ChelbaMSGBK13, Kwak10www, hermann2015teaching, attend2u:2017:CVPR}.
Genie also includes corpora of English text, both completely free-form, and specific to messages, social media captions and news articles.
This allows Genie to understand parameter values outside a closed knowledge base and provides a fallback for generic parameters.
%that do not have a more specific type.
Overall, Genie's database includes over 7.8 million distinct parameter values, of which 3 million are for free-form text parameters.
Genie expands the synthesized and paraphrase dataset by substituting parameters from user-supplied lists or its parameter databases. 
Finally, Genie also applies standard data augmentation techniques based on PPDB~\cite{ppdb} to the paraphrases.
%to generate the \textit{augmented} dataset.

\subsection{Combining Synthesized and Paraphrase Data}
Synthesized data are not only used to obtain paraphrases, but are also used as training data. 
Synthesized data provides variance in the space of programs, and enables the model to
learn type-based compositionality, while paraphrase data provides linguistic variety.

Developers can generate different sets of synthesized data according to the understandability of the functions or the presence of certain programming language features, such as compound commands, filters, timers, etc.  They can also control the size of each group by controlling the number of instantiations of each sentence with different parameters. 

\section{Neural Semantic Parsing Model}
\label{sec:nl-model}

Genie's semantic parser is based on Multi-Task Question Answering Network (MQAN), a previously-proposed model architecture that was found effective on a variety of NLP tasks~\cite{McCann2018decaNLP} such as Machine Translation, Summarization, Semantic Parsing, etc.
MQAN frames all NLP tasks as contextual question-answering. A single model can be trained on multiple tasks (\textit{multi-task training}~\cite{caruana1997mtl}), and MQAN uses the question to switch between tasks. In our semantic parsing task, the context is the natural language input from the user, and the answer is the corresponding program. The question is fixed because Genie does not use multi-task training, and has a single input sentence.

\iffalse
The MQAN model is an encoder-decoder architecture~\cite{sutskever2014sequence} that uses both recurrent and self-attentive~\cite{vaswani2017attention} layers.
\fi

\iftrue
\begin{figure}
\centering
\includegraphics[width=\linewidth,trim={0 2cm 0 2cm},clip]{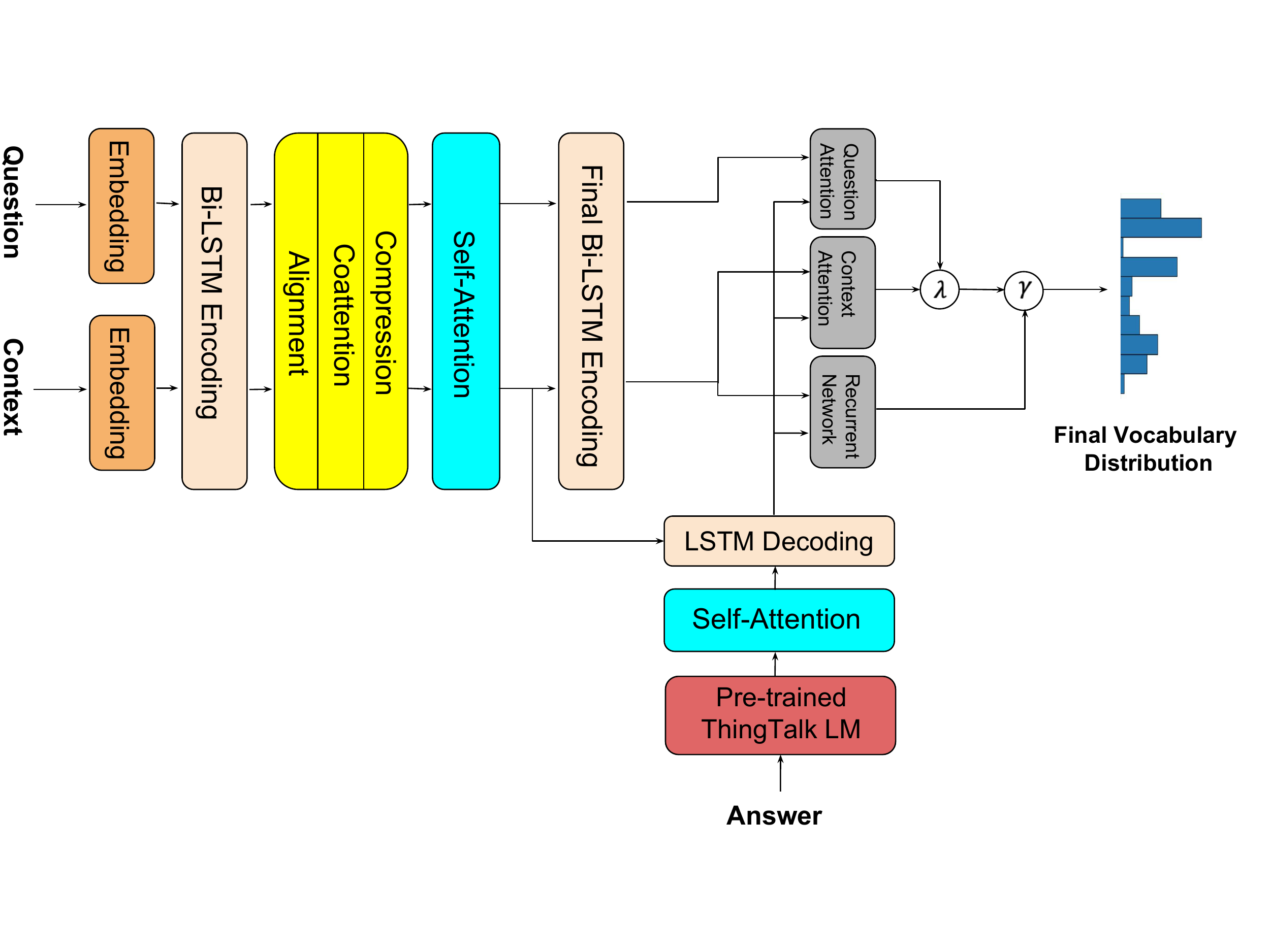}
\vspace{-1.5em}
\caption{Model architecture of Genie's semantic parser.}
\label{fig:mqan_model}
\end{figure}
\fi

\subsection{Model Description}
In MQAN both the encoder and decoder use a deep stack of recurrent, attentive and feed-forward layers to construct their input representations. In the encoder, each context and question are first embedded by concatenating word and character embeddings, and then fed to the encoder to construct the context and questions representations. 
%In the decoder, each answer is embedded by passing through a pre-trained language model layer.
The decoder uses a mixed pointer-generator architecture to predict the target program one token at a time; at each step, the decoder predicts a probability of either copying a token from the context or question, or generating one from a vocabulary, and then computes a distribution over input words and a distribution over vocabulary words. Two learnable scalar switches are then used to weight each distribution; the output is the token with the highest probability, which is then fed to the next step of the decoder, auto-regressively. The model is trained using token-level cross-entropy loss. We refer the readers to the original paper~\cite{McCann2018decaNLP} for more details.

\iffalse
\begin{small}
\vspace{-1em}
\begin{align*}
p(w_t) &= \gamma*p_v(w_t) + (1-\gamma)[\lambda*p_c(w_t) + (1-\lambda*p_q(w_t))]\\
\hat{w}_t &= \arg\max p(w_t)
\end{align*}
\vspace{-1em}
\end{small}

where $p_c(w_t)$, $p_q(w_t)$, and $p_v(w_t)$ are the probabilty distribiton over the set of words in context, question and external vocabulary respectively; $p(w_t)$ is the final probabilty distribution over the union of all the words. $\gamma$ and $\lambda$ indicate context-question and external vocabulary switch. $\hat{w}_t$ is the chosen word at time-step t.
\fi

\subsection{\thingtalk Language Model}

Genie applies a pre-trained recurrent language model (LM)~\cite{mikolov2010recurrent, DBLP:journals/corr/RamachandranLL16} to encode the answer (ThingTalk program) before feeding it to MQAN. The high-level model architecture is shown in Fig.~\ref{fig:mqan_model}. Previous work has shown that using supervised~\cite{mccann2017learned, DBLP:journals/corr/RamachandranLL16} and unsupervised~\cite{Peters:2018} pre-trained language models as word embedding can be effective, because it captures meanings and relations in context, and exposes the model to words and concepts outside the current task. The \thingtalk LM is trained on a large set of synthesized programs. This exposes the model to a much larger space of programs than the paraphrase set, without disrupting the balance of paraphrases and synthesized data in training.

\iffalse

\begin{small}
\vspace{-1em}
\begin{align*}
C_{emb} &= W_c[\text{Glove}(C)||\text{Char}(C)]\\
Q_{emb} &= W_q[\text{Glove}(Q)||\text{Char}(Q)]\\
A_{emb} &= W_a~\text{LM}(A)
\end{align*}
\vspace{-1em}
\end{small}

where $C_{emb}$, $Q_{emb}$, $A_{emb}$ are context, question, and answer embeddings. \text{Glove}, \text{Char}, and \text{LM}  denote Glove word embeddings, n-gram character embeddings, and pre-trained language model respectively.
\fi

\subsection{Hyperparameters and Training Details}
Genie uses the implementation of MQAN provided by decaNLP~\cite{McCann2018decaNLP}, an open-source library. Preprocessing for tokenization and argument identification was performed using the CoreNLP library~\cite{corenlp}, and input words are embedded using pre-trained 300-dimensional GloVe~\cite{pennington2014glove} and 100-dimensional character n-gram embeddings~\cite{hashimoto}. The decoder embedding uses 1-layer LSTM language model, provided by the floyhub open source library~\cite{floyhub}, and is pre-trained on a synthesized set containing 20,168,672 programs. Dropout~\cite{srivastava2014dropout} is applied between all layers. Hyperparameters were tuned on the validation set, which is also used for early stopping. All our models are trained to perform the same number of parameter updates (100,000), using the Adam optimizer~\cite{kingma2014adam}. Training takes about 10 hours on a single GPU machine (Nvidia V100).

\section{Experimentation}
\label{sec:eval}

\begin{table*}
\fontsize{8}{10}\selectfont
\centering
{%\setlength\tabcolsep{2pt}
\begin{tabular}{p{6cm}|p{5.75cm}|p{4.75cm}}
\toprule
{\bf Modification} & {\bf Example (before)} & {\bf Example (after)}\\
\hline
Replace second-person pronouns to first-person & Blink your light & Blink my light \\
\hline
Replace placeholders with specific values & ... set the temperature to \_\_\_° & ... set the temperature to 25 C \\
\hline
Append the device name if ambiguous otherwise & Let the team know when it rains & Let the team know when it rains on Slack \\
\hline
Remove UI-related explanation & Make your Hue Lights color loop with this button & Make your Hue Lights color loop \\
\hline
Replace under-specified parameters with real values & Message my partner when I leave work & Message John when I leave work \\
\bottomrule
\end{tabular}
}
\caption{IFTTT dataset cleanup rules.}
\label{table:ifttt-cleanup}
\vspace{-1.5em}
\end{table*}

% \TODO{WE NEED A MORE INFORMATIVE INTRO}
This section evaluates the performance of Genie on ThingTalk.
Previous neural models trained on paraphrase data have only been evaluated with paraphrase data~\cite{overnight}.
However, getting good accuracy on paraphrase data does not necessarily mean good accuracy on real data. 
Because acquiring real data is prohibitively expensive, our strategy is to create a high-quality training set based on synthesis and paraphrasing, and to validate and test our model with a small set of realistic data that mimics the real user input. 

In the following, we first describe our datasets. We evaluate the model on paraphrases whose programs are not represented in training. This measures the model's compositionality, which is important since real user input is unlikely to match any of the synthesized programs, due to the large program space.  Next, we evaluate Genie on the realistic data.  We compare Genie's training strategy against models trained with either just synthesized or paraphrase data, and perform an ablation study on language and model features. Finally, we analyze the errors and discuss the limitations of the methodology. 

Our experiments are performed on the set of Thingpedia skills available at the beginning of the study, which consists of 131 functions, 178 distinct parameters, and 44 skills.
Our evaluation metric is \textit{program accuracy}, 
which considers the result to be correct only if the output has the correct functions, parameters, joins, and filters.  This is equivalent to having the output match the canonicalized generated program exactly.  
To account for ambiguity, we manually annotate each sentence in the test sets with all programs that provide a valid interpretation.
The Genie toolkit and our datasets are available for download from our GitHub page\footnote{\url{https://github.com/stanford-oval/genie-toolkit}}.

\iffalse
but our ultimate goal is to achieve good performance on realistic data. 
Unfortunately, we can only afford to acquire realistic data for validation and test, and not for training.
This requires our training set to be representative of the real user input, without containing the actual data.
Moreover, the space of VAPL programs is so large that the user will not provide the same exact command that has been seen in training or even a paraphrase of it. Therefore, we need  our model to be compositional, and generalize to composition of functions, parameters and filters not seen in training.

We first evaluate compositionality, and we assess Genie's ability to generate a large and high quality training set, in comparison to our previous work.
To evaluate the representativeness of the training set, we evaluate Genie's full model for \TT on realistic data, and perform error analysis. Next, 
\fi

\iffalse
Our training set consists of synthesized data and paraphrase data. To achieve good accuracy on unseen realistic data, we need our data to be representative; this is obtained by writing templates that are informed from validation data. Additionally, we need our model to be robust and compositional, as to generalize to sentences not seen in training.

furthermore, because the domains tested are small, all logical forms that appear in testing also appear in training. 
Testing on paraphrase data is undesirable because paraphrases may not represent real usage data. 
\fi

\subsection{Evaluation Data}
\label{sec:evaluationdata}
We used three methods to gather data that mimics real usage: from the developers, from users shown a cheatsheat containing functions in Thingpedia, and IFTTT users. Because of the cost in acquiring such data, we managed to gather only 1820 sentences, partitioned into a 1480-sentence validation set and a 340-sentence test set.  

\iffalse
Furthermore, while paraphrase sentences acquired through Genie are correct and realistic sentences, they are not necessarily distributed in the same way as the real user inputs to a virtual assistant.
%, and testing on them would yield unnaturally high performance
To evaluate how Genie performs on more realistic inputs, 
\fi

\paragraph{Developer Data}
Almond has an online training interface %\footnote{\url{https://almond.stanford.edu/thingpedia/developers/train}}
that lets developers and contributors of Thingpedia write and annotate sentences.  
1024 sentences are collected from this interface from various developers, including the authors.
In addition, we manually write and annotate 157 sentences that we find useful for our own use.  
In total, we collect 1174 sentences, corresponding to 642 programs, which we refer to as the {\em developer data}. 

\paragraph{Cheatsheet Data}
Users are expected to discover Almond's functionality by scanning a cheatsheet containing phrases referring to all supported functions~\cite{almondwww17}. 
 %They can also choose phrases for specific devices interactively.
We collect the next set of data from crowdsource workers by first showing them phrases of 15 randomly sampled skills.  We then remove the cheatsheet and ask them to write compound commands that they would find useful, using the services they just saw. We annotate those sentences that can be implemented with the functions in Thingpedia. 
By removing the cheatsheet before writing commands, we prevent workers from copying the sentences verbatim and thus obtain more diverse and realistic sentences.
% We find this appropriate as in this work we target expert users who are familiar with the virtual assistant and the terms to use for each device. 
Through this technique, we collect 435 sentences, corresponding to 342 distinct programs.

\paragraph{IFTTT Data}
We use IFTTT as a totally independent source of sentences for test data.  
IFTTT allows users to construct applets, a subset of the ThingTalk grammar, using a graphical user interface.  Each applet comes with a natural language description.
We collect descriptions of the most popular applets supported by Thingpedia and manually annotate them. 

\iffalse
\TT program\footnote{Automatic translation from IFTTT is not possible, as the details of the program are not available. This is different from previous work using IFTTT~\cite{quirk15} as IFTTT has taken steps to prevent scraping.}. 
\fi
%We collect 1787 such descriptions from IFTTT, using the most popular commands of the services that are common between IFTTT and Thingpedia; of these, (~15)\% are supported, and they are manually annotated with the corresponding \TT program\footnote{Automatic translation from IFTTT is not possible, as the details of the program are not available.}. 

Because descriptions are not commands, they are often incomplete; e.g. virtual assistants are not expected to interpret a description like ``IG to FB'' to mean posting an Instagram photo on Facebook.  We adapt the complete descriptions using the rules shown in Table~\ref{table:ifttt-cleanup}.
In total, we collect 211 IFTTT sentences, corresponding to 154 programs.

%As these collected sentences are descriptions instead of commands, they are often template-style with slots for users to fill when enabling the applets. Some of them also miss important information of the command and rely on the GUI to specify the details. Therefore, we drop those descriptions that are not command-like or impossible to parse because of lack of information, and clean the rest based on 

We create a 1480-sentence \textit{validation set}, consisting of the entire developer data, 208 sentences from cheatsheet and 98 from IFTTT.  The remaining 227 cheatsheet sentences and 113 IFTTT sentences are used as our \textit{test set}; this set provides the final metric to evaluate the quality of the \TT parser that Genie produces.

\begin{figure}
\centering
\includegraphics[width=0.9\linewidth,trim={4.5cm 0 0.5cm 0},clip]{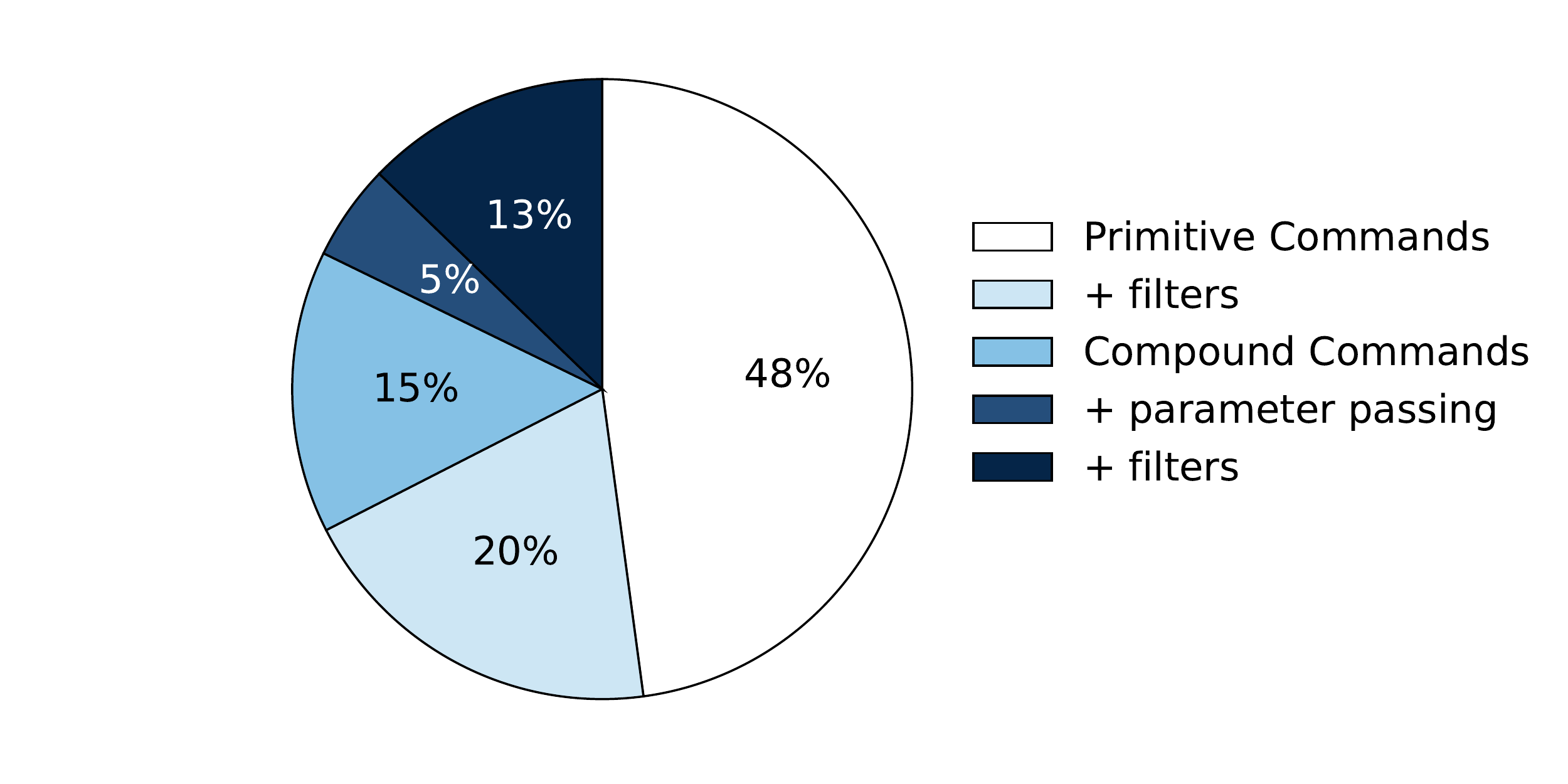}
\vspace{-1em}
\caption{Characteristics of the \TT training set (combining paraphrases and synthesized data).}
\label{fig:dataset_pie}
\vspace{-1em}
\end{figure}

\subsection{Evaluation}

\paragraph{Limitation of Paraphrase Tests}
Previous paraphrase-based semantic parsers are evaluated only with paraphrases similar to the training data~\cite{overnight}.  Furthermore, previous work synthesizes programs with just one construct template and one primitive template per function.  Using this methodology on the {\em original} ThingTalk, a dataset of 8,195 paraphrase sentences was collected.  After applying Genie's data augmentation on this dataset, the model achieves 95\% accuracy.  However, if we test the model on paraphrases of compound programs whose function combinations differ from those in training, the accuracy drops to 48\%.  This suggests that the dataset is too small to teach the model compositionality, even for paraphrases of synthesized data.  The accuracy drops to around 40\% with the realistic validation set.  This result shows that a high accuracy on a paraphrase test that matches programs in training is a poor indication of the quality of the model.  More importantly, we could not improve the result by collecting more paraphrases using this methodology, as the new paraphrases do not introduce much variation. 

\paragraph{Training Data Acquisition}

Templates help in improving the training data by allowing more varied sentences to be collected.  Our ThingTalk modification which combines triggers and retrievals as queries is also important; queries can be expressed as noun phrases, which can be inserted easily in richer construct templates.  We wrote 35 construct templates for primitive commands, 42 for compound commands, and 68 for filters and parameters. In addition, we also wrote 1119 primitive templates, which is 8.5 templates per function on average.

% We use Genie to generate a dataset with 2,869,390 synthesized sentences, 299,360 paraphrase sentences, and 183,035 augmented sentences.
Using the new templates, we synthesize 1,724,553 sentences, corresponding to 77,716 distinct programs, using a target size of 100,000 samples per grammar rule and a maximum depth of 5. With these settings, the synthesis takes about 25 minutes and uses around 23GBs of memory. The synthesized set is then sampled and paraphrased into 24,451 paraphrased sentences.
% Using Genie, we synthesized 1,724,553 sentences, which sample and paraphrase into 24,451 paraphrase sentences.
After PPDB augmentation and parameter expansion, the training set contains 3,649,222 sentences. Paraphrases with string parameters are expanded 30 times, other paraphrases 10 times, synthesized primitive commands 4 times, and other synthesized sentences only once. Overall, paraphrases comprise 19\% of the training set. %\TODO{Give numbers on syntehesized vs paraphrase data.}
%We use Genie to generate a dataset with 587,560 natural language and program pairs.
The characteristic of this combined dataset is shown in Fig.~\ref{fig:dataset_pie}; primitive commands are commands that use one function, while compound commands use two.
The training set contains 680,408 distinct programs, which include 4,710 unique combinations of Thingpedia functions.
%Note that the numbers in the table do not account for parameter expansion performed on the training set.
%Each command uses one or two functions from the library.

Sentences in the synthesized set use 770 distinct words; this number increases to 2,104 after paraphrasing and to 208,429 after PPDB augmentation and parameter expansion.
On average, each paraphrase introduces 38\% new words and 65\% new bigrams to a synthesized sentence.

%The Genie technique is iterative: after training a model, and evaluating on the development set, the developer refines the construct templates, and tune the proportion of paraphrases and synthesized sentences in the training set. The iteration continue until a satisfactory accuracy is obtained on the development set, or no more improvements are observed.

\begin{figure}
\includegraphics[width=0.85\linewidth]{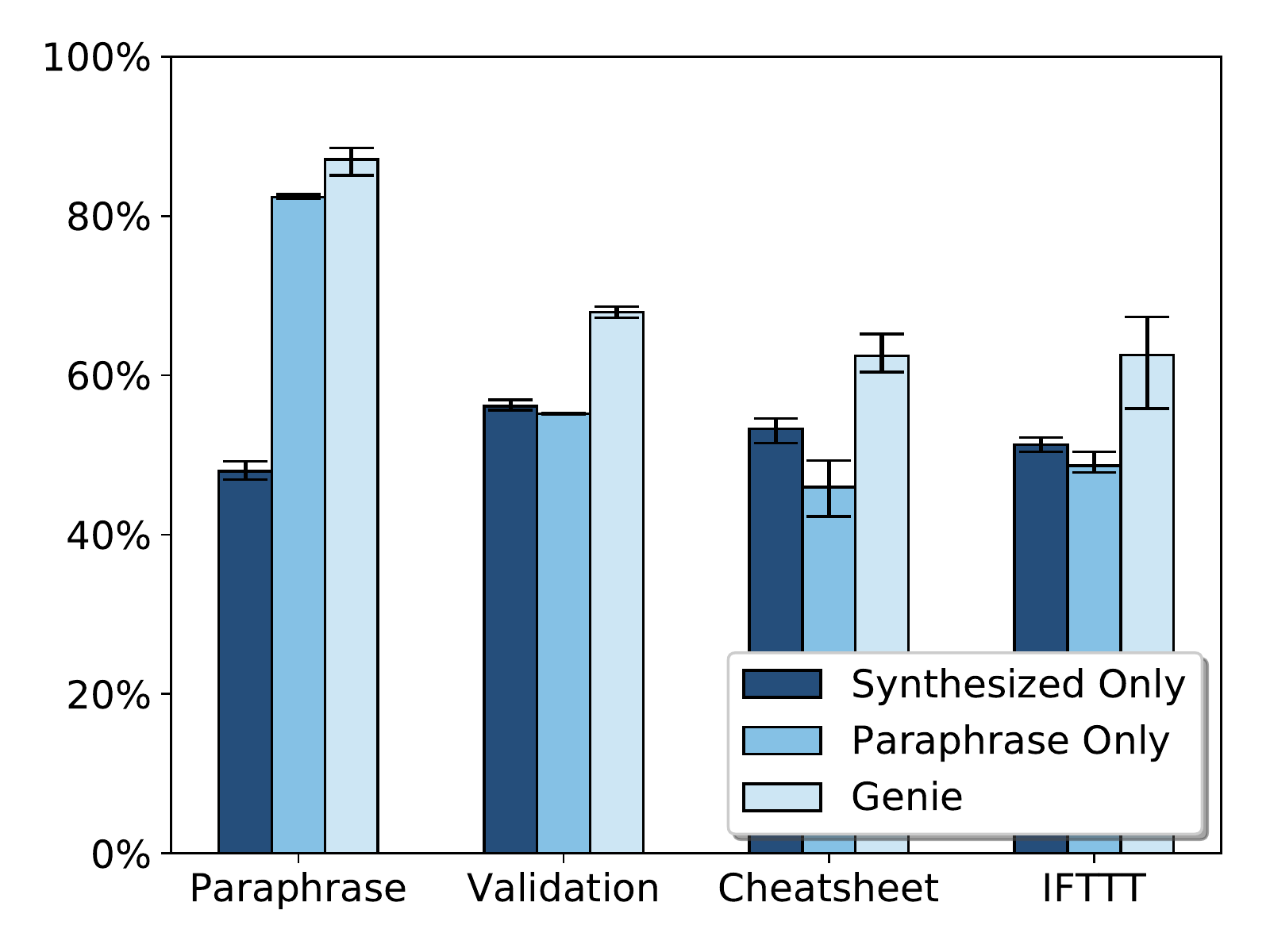}
\caption{Accuracy of the Genie model trained on just synthesized data, just paraphrase data, or with the Genie training strategy. Error bars indicate the range of results over 3 independently trained models.}
\label{fig:accuracy-all}
\vspace{-1em}
\end{figure}

\paragraph{Evaluation on Paraphrases with Untrained Programs}
Our paraphrase test set contains 1,274 sentences, corresponding to 600 programs and 149 pairs of functions. All test sentences are compound commands that use function combinations {\em not} appearing in training.
We show in Fig.~\ref{fig:accuracy-all} the average, minimum, and maximum of program accuracy obtained with 3 different training runs.  
On the paraphrase test set, Genie obtains an average program accuracy of 87\%, showing that Genie can 
generalize to compound paraphrase commands for programs not seen in training.  This improvement in generalization is due to both the increased size and variance in the training set.

\paragraph{Evaluation on Test Data}
%While evaluation on paraphrases is informative of the generalization power of the model, our goal is to achieve good performance on realistic data. Hence, we need to tune both the templates and the training data proportions on the validation set.  

%After a few rounds of tuning and template refinement, 
% Our first experimentation with the validation data returns around 40\% accuracy.   We apply the Genie methodology and synthesize more training data using templates created to match the patterns observed in the validation set; no paraphrases are obtained on such synthesized data. 

For the validation data described in Section~\ref{sec:evaluationdata}, Genie achieves an average of 68\% program accuracy. Of the 1480 sentences in the validation set, 272 sentences map to programs not in training; the rest uses different string parameters to programs that appear in the training set. Genie's accuracy is 30\% for the former and 77\% for the latter. This suggests that we need to improve on Genie's compositionality for sentences in the wild. 

When applied to the {\em test} data, Genie achieves full program accuracy of 62\%  on cheatsheet data and 63\% IFTTT commands, respectively. 
The difference in accuracy between the paraphrase test data and the realistic test sets underscores the limitation of paraphrase testing. 
Previous work on IFTTT to parse the high-level natural language descriptions of the rules can only achieve a 3\% program accuracy~\cite{quirk15}. 

%Our result shows, however, that if we can create a small validation set, we can write a few templates to boost the performance of the model on realistic test data. 

% Yet, even when we compare the function accuracy, the best reported result (an ensemble of 10 models, tested on the high quality subset of the test set) is 87.5\%~\cite{liu2016latent}. This result is lower than \TT, which confirms that it is not enough for the test set to be precise, if the training set is noisy.

%This underscores the difference, and the danger, of testing on paraphrase data, which can significantly over-estimate the performance of the model on realistic data. 
%Yet, Genie is able to achieve good accuracy on data that has no representation in the training set.  NOT TRUE AT ALL!  You created templates.  

% \paragraph{Error Analysis}
% \subsection{Error Analysis}

\iffalse
Function accuracy on validation data is 73\%.
Error analysis on the Genie model on the validation set reveals that the major cause of errors is wrong parameter names, filters and operators, which account for 43\% of the errors. 15\% of the errors are due to mistaking a primitive for a compound (or vice versa), 11\% of the errors are caused by an incorrect device, and 15\% by the incorrect function. Finally, 14\% of the errors are caused by producing a syntactically incorrect program, and less than 1\% of the errors are caused by the copying the wrong words from the input sentence.
\fi

\subsection{Synthesized and Paraphrase Training Strategy}
\label{sec:new-combinations}

%\TT is designed for an extensible Thingpedia; although it currently only has 213 primitives, it is intended to grow.  

%we create a new training/test split of the Paraphrase portion of the \TT dataset, such that none of the function combinations in the test sets are seen in the training sets. With this new split, the training set contains 2,941,858 sentences, the validation set 914 and the test set 1,017. All sentences in the validation and test set are paraphrased compound commands; each paraphrase appears only once in the test and validation sets.

The traditional methodology is to train with just paraphrased sentences.  As shown in Fig.~\ref{fig:accuracy-all}, training on paraphrase data alone delivers a program accuracy of 82\% on the paraphrase test, 55\% on the validation set, 46\% on cheatsheet test data, and 49\% on IFTTT test data.  This suggests that the smaller size of the paraphrase set, even after data augmentation, causes the model to overfit.  Adding synthesized data to training improves the accuracy across the board, and especially for the validation and real data test sets.  

On the other hand, training with synthesized data alone delivers a program accuracy of 48\% on the paraphrase test, 56\% on the validation set, 53\% on cheatsheet test data, and 51\% on IFTTT test data.  When compared to paraphrase data training, it performs poorly on the paraphrase test, but performs better on the cheatsheet and IFTTT test data.  

The combinination of using synthesized and paraphrase data works best. 
The synthesized data teaches the neural model many combinations not seen in the paraphrases, and the paraphrases teach the model natural language usage. 
Thus, synthesized data os not just useful as inputs to paraphrasing, but can expand the training dataset effectively and inexpensively.

\subsection{Evaluation of \VAPL and Model Features}
Here we perform an ablation study, where we remove a feature at a time from the design of Genie or \TT to evaluate its impact. 
We report, in Table~\ref{table:ablation}, results on three datasets: the paraphrase test set, the validation set, and those programs in the validation set that have new combinations of functions, filters, and parameters not seen in training.
We report the average across three training runs, along with the error representing the half range of results obtained.

\begin{table}\small
\centering
{\setlength\tabcolsep{2pt}
\begin{tabular}{lccc}
\toprule
 {\bf Model}          & {\bf Paraphrase}  & {\bf Validation} & {\bf New Program}               \\
\midrule
Genie                & $87.1 \pm 1.8$                    & $\textbf{67.9} \pm \textbf{0.7}$  & $29.9 \pm 3.2$                    \\
$-$ canonicalization & $80.0 \pm 1.3$                    & $63.2 \pm 0.9$                    & $21.9 \pm 0.9$                    \\
$-$ keyword param.   & $84.0 \pm 0.6$                    & $66.6 \pm 0.3$                    & $25.0 \pm 2.0$                    \\
$-$ type annotations & $86.9 \pm 3.6$                    & $67.5 \pm 0.6$                    & $\textbf{31.0} \pm \textbf{1.1}$                \\
$-$ param. expansion & $78.3 \pm 4.8$                    & $66.3 \pm 0.4$                    & $30.5 \pm 1.3$                    \\
$-$ decoder LM       & $\textbf{88.7} \pm \textbf{1.0}$  & $66.8 \pm 0.8$                    & $27.3 \pm 1.7$                    \\
\bottomrule
\end{tabular}

}
\caption{Accuracy results for the ablation study.\\
Each ``$-$'' row removes one feature independently.}
\vspace{-2em}
\label{table:ablation}
\end{table}

\paragraph{Canonicalization}
We evaluate canonicalization by training a model where keyword parameters are shuffled independently on each training example. (Note that programs are canonicalized during evaluation).  
Canonicalization is the most important \NLPL feature, improving the accuracy by 5 to 8\% across the three datasets. 

\paragraph{Keyword Parameters}
Replacing keyword parameter with positional parameters decreases performance by 3\% on paraphrases and 5\% on new programs in the validation set.  This suggests that keyword parameters improve generalization to new programs.

\paragraph{Type Annotations}
Type annotations are found to have no measurable effect on the accuracy, as the differences are within the margin of error. We postulate that keyword parameters are an effective substitute for type annotations, as they convey not just the type information but the semantics of the data as well.

\paragraph{Parameter Expansion}
Removing parameter expansion means that every sentence now appears only once in the training set.  This 
decreases performance by 9\% on paraphrases due to overfitting.  It has little effect on the validation set because only a small set of parameters are used. 

\paragraph{Pretrained Decoder Language Model} 
We compare our full model against one that uses a randomly initialized and jointly trained embedding matrix for the program tokens. Our proposal to augment the MQAN model with a pretrained language model improves the performance on new programs by 3\%, because the model is exposed to more programs during pretraining.  It lowers the performance for the paraphrase set slightly, however, because the basic model fits the paraphrase distribution well.

\subsection{Discussion}
Error analysis on the validation set reveals that Genie produces a syntactically correct and type-correct program for 96\% of the inputs, indicating that
the model can learn syntax and type information well.
Genie identifies correctly whether the input is a primitive or a compound with 91\% accuracy, and can identify the correct skills for 87\% of the inputs.
82\% of the generated programs use the correct functions; the latter metric corresponds to the \textit{function accuracy} introduced in previous work on IFTTT~\cite{quirk15}.
Finally, less than 1\% of the inputs have the correct functions, parameter names, and filters but copy the wrong parameter value from the input.

As observed on the validation set, the main source of errors is due to the difficulty of generalizing to programs not seen in training.
Because the model can generalize well on paraphrases, we believe this is not just a property of the neural model.
Real natural language can involve vocabulary that is inherently not compositional, such as ``autoreply'' or ``forward''; some of that vocabulary
can also be specific to a particular domain, like ``retweet''. It is necessary to learn the vocabulary from real data. 
The semantic parser generated by Genie may be used as beta software to get real user input, from which more templates can be derived.

\section{Case Studies of Genie}
\label{sec:case-studies}
We now apply Genie to three use cases: 
a sophisticated music playing skill,  an access control language modeled after \TT, and an extension to \TT for aggregates. We compare the Genie results with a {\em Baseline} modeled after the Wang et al. methodology~\cite{overnight}: training only with paraphrase data, no data augmentation, and no parameter expansion. 

\iffalse
We compare a model trained using Genie against a baseline trained with only paraphrase data and without any data augmentation or parameter expansion; this baseline closely matches the methodology introduced by Wang et al.~\cite{overnight}.

In these experiments, we consider extensions of the virtual assistant capabilities, and capabilities that do not exist in IFTTT or comparable commercial services; thus, we only evaluate on cheatsheet data.
\fi

\iffalse
We run three case studies and evaluate how Genie generalizes to different use cases and formal languages. 
We first evaluate Genie on a domain-specific music virtual assistant which uses Spotify APIs,
then we evaluate it on two extensions of \TT: {\em \TTa} and {\em \TACL}.
\TTa adds aggregation operators to \TT: an aggregation operator can be applied to any query function that returns a list and returns the sum, average, minimum or maximum value of one of the parameters.
\TACL defines access control policies for virtual assistant commands; it allows users to express for whom, what, when, where, and how \TT commands can be executed. 
\fi

\begin{figure}
\centering
\includegraphics[width=0.85\linewidth]{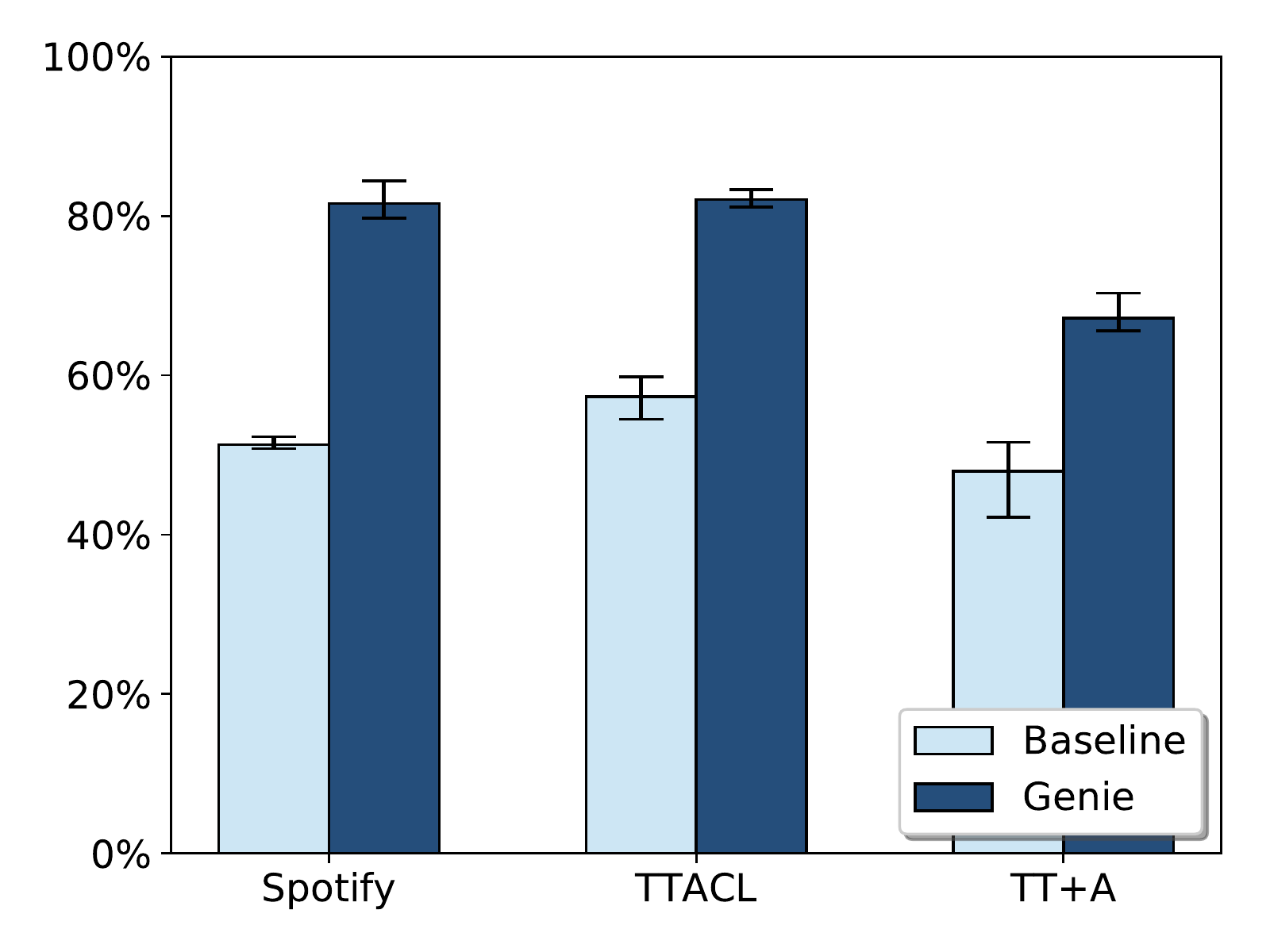}
\vspace{-1em}
\caption{Accuracy of the three case studies on cheatsheet test data. Baseline refers to a model trained with no synthesized data, no PPDB augmentation, and no parameter expansion. Error bars indicate the range of results over 3 independently trained models.}
\label{fig:acc-case-studies}
\vspace{-1em}
\end{figure}
\iffalse
 % Gabby Wright and Hemanth Kini, 
\fi
\subsection{A Comprehensive Spotify Skill}

The Spotify skill in Almond~\cite{spotifyskill} allows users to combine 15 queries and 17 actions in Spotify
in creative ways. 
%, by leveraging compositions of the public Spotify API, allows users to program their music player with personalization.
%Our first use case is to develop a Spotify skill service that supports combinations of Spotify primitives.  The Spotify skill was created . 
% 
%Users can combine the 15 queries and 17 actions in the skill in many interesting ways. 
For example,
users can ``add all songs faster than 500 bpm to the playlist dance dance revolution'', or ``wake me up at 8 am by playing wake me up inside by evanescence''. 

This skill illustrates the importance of Genie's ability to handle quote-free sentences.  In the original ThingTalk, a preprocessor replaces all the arguments, which must be quoted, with a \textsc{PARAM} token.  This would have replaced ``play `shake it off' '' and ``play `Taylor Swift' '' with the same input ``play $\texttt{PARAM}$`''.  However, these two sentences correspond to different API calls in Spotify because the former plays a given song and the latter plays songs by a given artist.  Genie accepts the sentences without quotes and uses machine learning to distinguish between songs and artists. 
Unlike in previous experiments, since the parameter value is meaningful in identifying the function, 
we use multiple instances of the same sentence with different parameters in the test sets.

The skill developers wrote 187 templates (5.8 per function on average) and collected 1,553 paraphrases. After parameter expansion, we obtain a dataset with 165,778 synthesized sentences and 217,258 paraphrases. %205,923 paraphrases, and 11,335 augmented sentences. %It contains 7,124 unique compound programs and 21,645 corresponding sentences.  \TODO{how is 21656 related  to all the 3 other numbers?}

We evaluate on two test sets: a paraphrase set of 684 paraphrase sentences, and a cheatsheet test set of 128 commands; the latter set contains 95 primitive commands and 33 compound commands.
We also keep 675 paraphrases in the validation set, and train the model with the rest of the data. 

%We also collect a cheatsheet dataset for Spotify commands; this set contains, corresponding to 95 primitive commands and 33 compound commands.
On paraphrases, Genie achieves a 98\% program accuracy, while the Baseline model achieves 86\%. On cheatsheet data, Genie achieves 82\% accuracy, an improvement of 31\% over the Baseline model (Fig.~\ref{fig:acc-case-studies}).  Parameter expansion is mainly responsible for the improvement because it is critical to identify the song or artist in the sentences.  %This suggests that Genie can be used by skill developers wishing to leverage compound commands. 

\iffalse
When trained with the full Genie model, the model achieves 89.5\% program accuracy, but only 84\% exact accuracy (with
parameters) 
The use of softmax loss, rather than max margin, increases the program accuracy to 91.7\%, and exact accuracy to 86.2\%.
Function accuracy is 95.1\% for the max-margin model, and 93.3\% for the softmax model.
As expect, the program and function accuracy is significantly higher for the domain-specific assistant; at the same time,
the exact accuracy does not increase significantly, possibly because the neural network must focus more on the specific
parameter values during training and cannot learn to copy all unknown words. Indeed, the exact accuracy of the retrieval baseline is
just 15.8\%, despite a program accuracy of 53.4\%, because the hardness of this task is in the parameters.
\fi

\subsection{\TT Access Control Language}

We next use Genie for TACL, an access control policy language that lets users describe who, what, when, where, and how their data can be shared~\cite{commaimwut18}.  A policy consists of the person requesting access and a \TT command.  For example, the policy ``my secretary is allowed to see my work emails'' is expressed as:

%For our next use case in this paper, we use Genie to add access control to \TT to create \TTACL. 
%We add a source, the person requesting access, to primitive \TT commands; the source can be used as parameters as well as filters.  For example, the command ``my secretary is allowed to see my work emails'' is expressed as:

\begin{small}
\begin{tabbing}
123\=123\=\kill
\>$\sigma = \text{``secretary''} : \texttt{now}$\\
\>$\Rightarrow$\>$@\text{com.gmail.inbox}\texttt{ filter } \textit{labels} \texttt{ contains } \text{``work''}$\\
\>$\Rightarrow$\>$\texttt{notify}$
\end{tabbing}
\end{small}

The previously reported semantic parser handles only primitive TACL policies, whose formal grammar is shown in Fig.~\ref{fig:tacl-grammar}.  All parameters are expected to be quoted, which renders the model impractical with spoken commands.  An accuracy of 74\% was achieved on a dataset consisting of 4,742 paraphrased policy commands; the number includes both training and test.

\begin{figure}
\fontsize{8}{10}\selectfont
\begin{tabbing}
\=123\=12345678901234567890\=\kill
\iffalse
\> Second-party \TT: \\
\>\>Program $\pi$: \> $\texttt{source}=\sigma~~\texttt{executor}=\epsilon~~\texttt{monitor }q\Rightarrow\texttt{notify}~~\vert~~\texttt{now}\Rightarrow q\Rightarrow\texttt{notify}~~\vert~~\texttt{now}\Rightarrow a;$\\
\>\>Source $\sigma$: \> $\textsc{self}~~\vert~~v$\\
\>\>Executor $\epsilon$: \> $\textsc{self}~~\vert~~v$\\\\
\> \TACL:\\
\fi
%\>\>Policy $\hat{\pi}$: \>$\hat{p_\sigma} : \left[\texttt{monitor }\hat{q}\Rightarrow\texttt{notify}~~\vert~~\texttt{now}\Rightarrow \hat{q}\Rightarrow\texttt{notify}~~\right.$\\
%\>\>\>$\left.\vert~~\texttt{now}\Rightarrow \hat{q}\right]$\\

\>\>Policy $\hat{\pi}$: \>$\hat{p_\sigma} : \left[\texttt{now}\Rightarrow \hat{q}\Rightarrow\texttt{notify}~~\vert~~\texttt{now}\Rightarrow \hat{a}\right]$\\
\>\>Query $\hat{q}$: \>$f~~\texttt{filter}~~p$\\
\>\>Action $\hat{a}$: \>$f~~\texttt{filter}~~p$\\
\>\>Source predicate $\hat{p_\sigma}$: \> $\texttt{true}~~\vert~~\texttt{false}~~\vert~~\texttt{!}p~~\vert~~p\texttt{ \&\& }p~~\vert~~
p~\texttt{||}~p~~\vert$\\
        \>\>\>$\sigma~\textit{operator}~v$
\end{tabbing}
\vspace{-1em}
\caption{The formal grammar of the primitive subset of TACL.
Rules that are identical to \TT are omitted.}
\vspace{-1em}
\label{fig:tacl-grammar}
\end{figure}

%We refine \TACL to be compatible with the more powerful \TT.
%For comparison with prior work, we first evaluate on paraphrases. 

Our first experiment reuses the same dataset above, but with quotes removed.  From 6 construct templates, Genie synthesizes 432,511 policies, and combines them with the existing dataset to form a total training set of 543,566 policies, after augmentation.
We split the dataset into 526,322 sentences for training, 701 for validation, and 702 for testing; the test consists exclusively of paraphrases unique to the whole set, even when ignoring the parameter values. %The training set was expanded to 318,361 sentences by replacing parameters with sampled values from the parameter lists supplied with corresponding library functions.
On this set, Genie achieves a high accuracy of 96\%.

For a more realistic evalution, we create a test set of 132 policy sentences, collected using the cheatsheet technique. We reuse the same cheatsheet as the main \TT experiment; instructions are modified to elicit access control policies rather than commands. On this set, Genie achieves an accuracy of 82\%, with a 25\% improvement over the Baseline model.
This shows that the data augmentation technique in Genie is effective in improving the accuracy, even if the paraphrase portion of the training set is relatively small.
The high accuracy is likely due to the limited scope of policy sentences: the number of meaningful primitive policies is relatively small compared to the full Thingpedia.

%they are only primitives, and the space of personal accounts that can be meaningfully shared is a significantly smaller set than the full 

\subsection{Adding Aggregation to \TT}

One of our goals in building Genie is to grow virtual assistants' ability to understand more commands.  Our final case study adds aggregation to \TT, i.e. finding min, max, sum, average, etc.  Our new language \TTa extends \TT with the following grammar:

\begin{small}
\begin{tabbing}
12\=123456789\=\kill
\>Query $q$: \>$\texttt{agg}~~\left[\texttt{max}~\vert~\texttt{min}~\vert~\texttt{sum}~\vert~\texttt{avg}\right]
~~\textit{pn}~~\texttt{of}~~(q)~~\vert$\\
        \>\>$\texttt{agg}~~\texttt{count}~~\texttt{of}~~(q)$
\end{tabbing}
\end{small}

Users can compute aggregations over results of queries, such as ``find the total size of a folder’’, which translates to

\begin{small}
\begin{tabbing}
123\=1234\=123\=\kill
\>$\texttt{now}$ \> $\Rightarrow \texttt{agg}~~\texttt{sum}~~\textit{file\_size}~~\texttt{of}~~\left(@\text{com.dropbox.list\_folder}()\right)$\\
\>\>$\Rightarrow$ \>$\texttt{notify}$
\end{tabbing}
\end{small}

While this query can be used as a clause in a compound command, we only test the neural network with aggregation on primitive queries.  
We wrote 6 templates for this language.

The \TT skill library currently has 4 query functions that return lists of numeric results, and 20 more that returns lists in general (on which the count operator can be applied).  We synthesize 82,875 sentences and collect 2,421 paraphrases of aggregation commands. For training, we add to the full \TT dataset 270,035 aggregation sentences: 23,611 are expanded paraphrases and the rest are synthesized. The accuracy on the paraphrase test set is close to 100\% for both Genie and the Baseline; this is due to the limited set of possible aggregation commands and the fact that programs in the paraphrase test set also appear in the training set.

We then test on a small set of 64 aggregation commands, obtained using the cheatsheet method. In this experiment, the cheatsheet is restricted to only queries where aggregation is possible. We note that the cheatsheet, in particular, does not show the output parameters of each API, so crowdsource workers guess which parameters are available to aggregate based on their knowledge of the function; this choice makes the data collection more challenging, but improves the realism of inputs because workers are less biased. On this set, Genie achieves a program accuracy of 67\% without any iteration on templates
(Fig.~\ref{fig:acc-case-studies}), an improvement of 19\% over the Baseline. This accuracy is in line with the general result on ThingTalk commands, and suggests that Genie can support extending the language ability of virtual assistants effectively.

\section{Related Work}
\label{sec:related}

\iffalse
{\bf Almond.}
The original publication of the Almond virtual assistant and ThingTalk language \cite{almondwww17} used classic semantic parsing to understand compound commands, based on the SEMPRE algorithm~\cite{pasupat2015compositional, overnight}, and reported an accuracy of 71\% for primitive commands and 51\% for compounds.
% The dataset was never released, so it is not possible to compare directly.
As discussed in Section~\ref{sec:new-combinations}, we improve our previous result by refining the ThingTalk language, and by collecting an improved, larger and more varied dataset.
% Yet, we believe the results obtained by Genie on the revised \TT language, are sufficient to show the improvement.
% We also note that both the ThingTalk language and dataset, as well as the Thingpedia repository have changed since the results were published.
\fi

\paragraph{Alexa}
The Alexa assistant is based on the Alexa Meaning Representation Language (AMRL)~\cite{kollar2018alexa}, a language they designed to support semantic parsing of Alexa commands.

AMRL models natural language closely: for example, the sentence ``find the sharks game and find me a restaurant near it''
would have a different representation than ``find me a restaurant near the sharks game''~\cite{kollar2018alexa}.
 In \TT, both sentences would have the same executable representation, which enables paraphrasers to switch from one to the other. 
%This is tied to the fact that properties (parameter names),
%types (entity mentions, operator values) must be explicitly aligned to a span of words in the input using inside-outside-beginning tagging~\cite{perera2018multi}. In \TT, both sentences would have the same normalized representation, which enables paraphrasers to switch from one to the other freely.

AMRL has been developed on a closed ontology of 93 actions and 60 intents, using a 
dataset of sentences manually annotated by experts (not released publicly). The best accuracy reported on this dataset is 77\%~\cite{perera2018multi}.
%Their model is also specifically tuned for spoken language understanding (joint intent classification and slot tagging) and it is not clear how well it fares on
%complex commands with multiple intents or operators.

AMRL is not available to third-party developers. Third-party skills have access to a joint intent-classification and slot-tagging model~\cite{goyal2018fast},
which is equivalent to a single \TT action. Free-form text parameters are further limited to one per sentence and must use a templatized
carrier phrase. The full power of \TT and Genie is instead available to all contributors to the library.

\paragraph{IFTTT}
If-This-Then-That~\cite{ifttt} is a service that allows users to combine services into trigger-action rules.
Previous work~\cite{quirk15} attempted to translate the English description of IFTTT rules into executable code.
Their method, and successive work using the IFTTT dataset~\cite{quirk16, dong2016language, liu2016latent, yin2017syntactic, alvarez2016tree,2018arXiv180806740Y}, showed moderate success in identifying the correct functions on a filtered set of unambiguous sentences but failed to identify the full programs with parameters.
They found that the descriptions are too high-level and are not precise commands. For this reason, the IFTTT dataset is unsuitable to train a semantic parser for a virtual assistant.%, and it is unclear that any improvement on the IFTTT dataset would transfer to more precise datasets, where exact match accuracy is the goal.

\paragraph{Data Acquisition and Augmentation}
Wang et al. propose to use paraphrasing technique to acquire data for semantic parsing;
they sample canonical sentences from a grammar, crowdsource paraphrases them and then use the paraphrases as training data~\cite{overnight}.
Su et al.~\cite{su2017building} explore different sampling methods to acquire paraphrases for Web API.
They focus on 2 APIs; our work explores a more general setting of 44 skills.

Previous work ~\cite{jia2016data} has proposed the use of a grammar of natural language for data augmentation. Their work supports data augmentation with power close to Genie's parameter expansion, but it does so with a grammar that is automatically inferred from natural language.
Hence, their work requires an initial dataset to infer the grammar, and cannot be used to bootstrap a new formal language from scratch.
 %Future work will investigate combining Genie with their technique, to offer a set of candidate constructs to the developer during template design and refinement.

A different line of work by Kang et al.~\cite{DBLP:journals/corr/abs-1809-02305} also considered the use of generative models to expand the training set; this is shown to increase accuracy. Kang et al. focus on joint intent recognition and slot filling. It is not clear how well their technique would generalize to the more complex problem of semantic parsing.

\paragraph{Semantic Parsing}
Genie's model is based upon previous work in semantic parsing, which was used for queries~\cite{zelle1994inducing, zelle1996learning, tang2001using, zettlemoyer2005learning, wong2007learning, berant2014semantic, pasupat2015compositional, overnight, xiao2016sequence, zhong2017seq2sql, xu2017sqlnet, iyer2017learning}, instructions to robotic agents~\cite{kate2005learning, kate2006using, wong2006learning, chen2011learning}, and trading card games~\cite{ling2016latent, yin2017syntactic, rabinovich2017abstract}.

Full SQL does not admit a canonical form, because query equivalence is undecidable~\cite{trakhtenbrot1950impossibility, chu2017cosette}, so previous work on database queries have targeted restricted but useful subsets~\cite{zhong2017seq2sql}.
%For the same reason, previous efforts to generate Python have focused exclusively on pseudo-code
%readouts of code snippets~\cite{yin2017syntactic}, or have obtained no exact accuracy and resorted
%to evaluating the BLEU score~\cite{rabinovich2017abstract}, which has no significance for a programming language.
\TT instead is designed to have a canonical form.

The state-of-the-art algorithm is sequence-to-sequence with attention~\cite{sutskever2014sequence, bahdanau2014neural, dong2016language}, optionally extended with a copying mechanism~\cite{jia2016data}, grammar structure~\cite{yin2017syntactic, rabinovich2017abstract}, and tree-based decoding~\cite{alvarez2016tree}. In our experiments, we found that the use of grammar structure provided no additional benefit.

\section{Conclusion}
\label{sec:conclusion2}

Virtual assistants can greatly simplify and enhance our lives. Yet, building a virtual assistant that supports novel capabilities is challenging because of a lack of annotated natural language training data. 
Commercial efforts use formal representations motivated by natural languages and labor-intensive manual annotation~\cite{kollar2018alexa}.
This is not scalable to the expected growth of virtual assistants. 

We advocate using a formal VAPL language to represent the capability of a virtual assistant and use a neural semantic parser to directly translate user input into executable code.  A previous attempt of this approach failed to create a good parser~\cite{almondwww17}.  This paper shows it is necessary to design the VAPL language in tandem with a more sophisticated data acquisition methodology. 
 
%However, our previous attempt to build Almond applied data acquisition technology that was state-of-the-art at the time, and yet struggled significantly to obtain high quality data~\cite{almondwww17}. We found a need to design the formal representation in tandem with the data acquisition methodology.

We propose a methodology and a toolkit, Genie, to create semantic parsers for new virtual assistant capabilities.  Developers only need to acquire a small set of manually annotated realistic data to use as validation.  They can get a high-quality training set by writing construct and primitive templates. Genie uses the templates to synthesize data, crowdsources paraphrases for a sample, and augments the data with large parameter datasets.  Developers can refine the templates iteratively to improve the quality of the training set and the resulting model.

We identify generally-applicable design principles that make \VAPL languages amenable to natural language translation.  By applying Genie to our revised \TT language, we obtain an accuracy of 62\% on realistic data. Our work is the first virtual assistant to support compound commands with unquoted free-form parameters. Previous work either required quotes~\cite{almondwww17} or obtained only 3\% accuracy~\cite{quirk15}. 

Finally, we show that Genie can be applied to three use cases in different domains; our methodology improves the accuracy between 19\% and 33\% compared to the previous state of the art. This suggests that Genie can be used to bootstrap new virtual assistant capabilities in a cost-effective manner.

Genie is developed as a part of the open-source Almond project~\cite{Almondwebsite}. Genie, our data sets, and our neural semantic parser, which we call LUInet (Linguistic User Interface network), are all freely available. Developers can use Genie to create cost-effective semantic parsers for their own domains. By collecting contributions in VAPL constructs, Thingpedia entries, and natural language sentences from developers in different domains, we can potentially grow LUInet to be the best and publicly available parser for virtual assistants.

\begin{acks}
We thank Rakesh Ramesh for his contributions to a previous version of Genie, and Hemanth Kini and Gabby Wright for their Spotify skill.
Finally, we thank the anonymous reviewers and the shepherd for their suggestions. 

This work is supported in part by the \grantsponsor{NSF}{National Science Foundation}{https://www.nsf.gov/awardsearch/showAward?AWD_ID=1900638&HistoricalAwards=false} under Grant No.~\grantnum{nsf}{1900638} and the Stanford MobiSocial Laboratory, sponsored by AVG, Google, HTC, Hitachi, ING Direct, Nokia, Samsung, Sony Ericsson, and UST Global.
\end{acks}

\bibliographystyle{ACM-Reference-Format}
\balance
\bibliography{paper}

\end{document}